%% file: main.tex
\pdfoutput=1
\documentclass[conference, 10pt]{IEEEtran}
\IEEEoverridecommandlockouts
\usepackage{cite}
\usepackage{amsmath,amssymb,amsfonts}
\usepackage{graphicx}
\usepackage{textcomp}
\usepackage{xcolor}
\usepackage{bm}
\usepackage{cases}
\usepackage[skip=1pt, belowskip=0pt]{caption}
\usepackage{subcaption} 
\usepackage[ruled,vlined,linesnumbered]{algorithm2e} 
    \usepackage{setspace} 
\usepackage[makeroom]{cancel} 
\usepackage{amsthm}
\usepackage{enumitem}  
\usepackage{multirow}
\usepackage[normalem]{ulem} 

    \usepackage{physics}
    \usepackage{amsmath}
    \usepackage{tikz}
    \usepackage{mathdots}
    \usepackage{yhmath}
    \usepackage{cancel}
    \usepackage{color}
    \usepackage{siunitx}
    \usepackage{array}
    \usepackage{multirow}
    \usepackage{amssymb}
    \usepackage{gensymb}
    \usepackage{tabularx}
    \usepackage{extarrows}
    \usepackage{booktabs}
    \usetikzlibrary{fadings}
    \usetikzlibrary{patterns}
    \usetikzlibrary{shadows.blur}
    \usetikzlibrary{shapes}
    
\makeatletter
\def\BibTeX{{\rm B\kern-.05em{\sc i\kern-.025em b}\kern-.08em
    T\kern-.1667em\lower.7ex\hbox{E}\kern-.125emX}}
    
\allowdisplaybreaks
    
\input{macros}

\begin{document}
    
    \bstctlcite{IEEEexample:BSTcontrol}

     \title{Distributed Continual Learning with \cocoa{} in High-dimensional Linear Regression}
    \author{
        \IEEEauthorblockN{
            Martin Hellkvist, 
            Ay\c ca \"Oz\c celikkale, 
            Anders Ahl\'{e}n
        }
        \thanks{
                The authors are with the Department of Electrical Engineering at Uppsala University, Uppsala, Sweden
                (e-mail: \{Martin.Hellkvist, Ayca.Ozcelikkale, Anders.Ahlen\}@angstrom.uu.se).
            }
    }
    
    \maketitle

\input{abstract}


    \input{my_cocoa}

    \input{definitions_theorems_etc}
    \input{introduction}

    \input{problem_statement}

    \input{main_results}
    \input{numerical_results}

    \input{conclusions}

    \input{appendix}

    \kern-0.25em
    
    \bstctlcite{IEEEexample:BSTcontrol}
    \bibliographystyle{IEEEtran}
    \bibliography{ref}

\end{document}

%% file: macros.tex
\newcommand{\Rbb}{\mathbb{R}}
\newcommand{\inrbb}[1]{\in\Rbb^{#1}}

\newcommand{\Nbb}{\mathbb{N}}
 
\mathchardef\myhyphen="2D

\def\qed{\hfill $\square$}

\newcommand{\Hmatt}[2]{\Hmat^{\mkern-1mu \{ \mkern-2mu #2 \mkern-2mu \}}_{\mkern-1.5mu #1}}
\newcommand{\Qmatt}[3]{\Qmat^{\mkern-1mu \{ \mkern-2mu #3 \mkern-2mu \}}_{\mkern-1mu #1 \mkern-2.5mu,\mkern1mu #2}}
\newcommand{\Rmatt}[2]{\Rmat^{\{ \mkern-2mu #2 \mkern-2mu \}}_{\mkern-0mu #1}}
\newcommand{\tauplusone}{\tau \mkern-1.5mu + \mkern-1.5mu 1}

\newcommand{\Tplusone}{T \mkern-1.5mu + \mkern-1.5mu 1}

\newcommand{\Amat}{\matr{A}}

\newcommand{\wbar}{\Bar{\wvec}}

\newcommand{\Abar}{\bar{\Amat}}

\newcommand{\Ltrain}[1]{\ell_{#1}}
\newcommand{\Forgetting}[1]{f_{#1}}
\newcommand{\Generalization}[1]{g_{#1}}
\newcommand{\traindata}[1]{\Dc_{#1}}
\newcommand{\offlinesoln}{\hat{\wvec}_{\text{LS}}}

\newcommand{\what}{\hat{\wvec}}

\newcommand{\cocoa}{\mbox{\textsc{CoCoA}}}
\newcommand{\cola}{\textsc{CoLa}}

\newcommand{\subproblemp}{\sigma'}

\newcommand{\aggregationp}{\bar \varphi}
\newcommand{\pmax}{p_{\max{}}}
\newcommand{\pmin}{p_{\min{}}}

\newcommand{\matr}[1]{\bm{#1}}

\newcommand{\Hmat}{\matr{H}}

\newcommand{\Mmat}{\matr{M}}
\newcommand{\Pmat}{\matr{P}}
\newcommand{\Qmat}{\matr{Q}}
\newcommand{\Rmat}{\matr{R}}

\newcommand{\Sigmamat}{\matr{\Sigma}}

\newcommand{\avec}{\matr{a}}

\newcommand{\qvec}{\matr{q}}
\newcommand{\svec}{\matr{s}}
\newcommand{\uvec}{\matr{u}}
\newcommand{\vvec}{\matr{v}}
\newcommand{\vbar}{\bar{\matr{v}}}
\newcommand{\wvec}{\matr{w}}
\newcommand{\xvec}{\matr{x}}

\newcommand{\yvec}{\matr{y}}
\newcommand{\zvec}{\matr{z}}

\newcommand{\muvec}{\matr{\mu}}

\newcommand{\zetavec}{\matr{\zeta}}

\newcommand{\Imat}{\matr{I}}

\renewcommand{\rank}{\text{Rank}}

\newcommand{\eye}[1]{\Imat_{#1}}

\DeclareMathOperator{\Ebb}{\mathbb{E}}
\newcommand{\eunder}[1]{\underset{#1}{\Ebb}}

\DeclareMathAlphabet{\mymathbb}{U}{BOONDOX-ds}{m}{n}
\newcommand{\onebb}{\mymathbb{1}}

\newcommand{\diag}[3]{\text{diag}\!\left\{#1\right\}_{\!#2}^{\!#3}}
\newcommand{\diagt}[3]{\text{diag}\! #1 _{\!#2}^{\!#3}}

\newcommand{\T}{^\intercal} 
\newcommand{\p}{^+}

\let\oldin\in
\renewcommand{\in}{{\,\oldin\,}}

\let\oldnotin\notin
\renewcommand{\notin}{{\,\oldnotin\,}}

\newcommand{\krange}{{k=1,\,\dots,\,K}}
\newcommand{\trange}{{t=1,\,\dots,\,T}}

\long\def\/*#1*/{}

\newcommand{\Dc }{\mathcal{D}}

\newcommand{\Nc }{\mathcal{N}}

\newcommand{\Wc}{\mathcal{W}}

\definecolor{myblue}{rgb}{0.3328, 0.3539, 0.7758}
\definecolor{myblue2}{rgb}{0.0328, 0.0539, 0.4758}
\definecolor{mygreen2}{rgb}{ 0.0328 0.4758 0.0539} 
\definecolor{mygreen3}{rgb}{ 0.0328 0.1758 0.0539} 
\definecolor{myred}{rgb}{0.4758, 0.0328, 0.0539}
\definecolor{myred2}{rgb}{0.75, 0.0328, 0.0539}

\newcommand{\balert}[1]{{\color{myblue}{#1}}}

\newcommand{\galert}[1]{{\color{myblue}{#1}}}

\renewcommand{\balert}[1]{{{#1}}}       
        
\renewcommand{\galert}[1]{{{#1}}}

%% file: abstract.tex
\begin{abstract}
    We consider estimation under scenarios where the  signals of interest exhibit change of characteristics over time.
    In particular,
    we consider the continual learning problem where different tasks, 
    e.g., data with different distributions,
    arrive sequentially and the aim is to perform well on the newly arrived task without performance degradation on the previously seen tasks. 
    In contrast to the continual learning literature focusing on the centralized setting, we investigate the problem from a  distributed estimation perspective. We consider the well-established distributed learning algorithm \cocoa{}, \balert{which distributes the model parameters and the corresponding features over the network}.
    We provide exact analytical characterization for the generalization error of \cocoa{} under continual learning \balert{for linear regression} in a range of scenarios, \balert{where overparameterization is of particular interest.}  
    These analytical results characterize 
    how the generalization error depends on the network structure,
    the task similarity and the number of tasks,
    and show how these dependencies are intertwined.
    In particular,
    our results show that 
    the generalization error can be significantly reduced by adjusting the network size,
    where the most favorable network size depends on task similarity and the number of tasks. 
    We present numerical results verifying the theoretical analysis and illustrate the continual learning performance of \cocoa{} with a digit classification task.  
\end{abstract}

%% file: my_cocoa.tex
\newcommand{\mycocoa}{
    \caption{\cocoa{} for task $t$}
    \label{alg:continual_cocoa} 
    
    \SetKw{KwEnd}{end}
	\textbf{Input}:
	Training data $\left(\yvec_t, 
        \Amat_t=[\Amat_{t,[1]}, \, \dots,\, \Amat_{t,[K]}]\right)$,
        previous estimate $\what_{t-1}^{(T_c)}$,
        number of iterations $T_c$ for \cocoa{} to run per task.
    
    \textbf{Initialize}:	
    $\what_{t}^{(0)}=\what_{t-1}^{(T_c)}$,    
    
    $\vvec_{t,[k]}^{(0)} = K \Amat_{t,[k]} \what_{t,[k]}^{(0)},\, k=1,\,\dots,\,K$.
    
    \setstretch{1.05}
    \For{$i=1,\,\dots,\,T_c$}{
        $\vbar_t^{(i)} = \frac{1}{K}\sum_{k=1}^K\vvec_{t,[k]}^{(i-1)}$
        
        \For{$k \in \{1,\,2,\,\dots,\,K\}$}{
        
            $\Delta \what_{t,[k]}^{(i)} 
                = \frac{1}{K}\Amat_{t,[k]}\p
                \left(\yvec_t - \vbar_t^{(i)}\right)$
            
            $\what_{t,[k]}^{(i)}  = \what_{t,[k]}^{(i-1)} + \Delta \what_{t,[k]}^{(i)} $
            
            $\vvec_{t,[k]}^{(i)} = \vbar_t^{(i)} + K\Amat_{t,[k]}\Delta \what_{t,[k]}^{(i)}$
        }
    }
    
    \textbf{Output: } $\what_{t}^{(T_c)}$
}

%% file: definitions_theorems_etc.tex
\newtheorem{thm}{\bf{Theorem}}
\newtheorem{cor}{\bf{Corollary}}
\newtheorem{lem}{\bf{Lemma}}
\newtheorem{prop}{\bf{Proposition}}
\newtheorem{rem}{\bf{Remark}}

\theoremstyle{remark} 
\newtheorem{defn}{\bf{Definition}}[section]
\newtheorem{ex}{\bf{Example}}
\newtheorem{myexp}{\bf{Experiment}}
\newtheorem{asmptn}{\bf{Assumption}}
\newtheorem{mymodel}{\bf{Task Model}}

\newenvironment{assumption}
{\par\noindent \asmptn \begin{itshape}\noindent}
{\end{itshape} \vspace{0pt}}

\newenvironment{theorem}
{\par\noindent \thm \begin{itshape}\noindent}
{\end{itshape} \vspace{3pt}}

\newenvironment{lemma}
{\par\noindent  \lem \begin{itshape}\noindent}
{\end{itshape} \vspace{3pt}}

\newenvironment{corollary}
{\par\noindent  \cor \begin{itshape}\noindent}
{\end{itshape} \vspace{3pt}
}

\newenvironment{remark}
{\par\noindent \rem \begin{itshape}\noindent}
{\end{itshape} \vspace{3pt}}

\newenvironment{definition}
{\par\noindent \defn \begin{itshape}\noindent}
{\end{itshape}}

\newenvironment{experiment}
{\vspace{3pt} \par\noindent \myexp \begin{itshape} \noindent}
{\end{itshape}\vspace{6pt}}

\newenvironment{example}
{\vspace{2pt} \par\noindent \ex} 
{\vspace{2pt}}

\newenvironment{model}
{\vspace{2pt} \par\noindent \mymodel} 
{\vspace{2pt}}

%% file: introduction.tex
\section{Introduction}\label{sec:introduction}

When presented with a stream of data,
continual learning \cite{Parisi_continual_2019, Kirkpatrick_catastrophic_2017} is the act of learning from new data while not forgetting what was learnt previously.
New data can, for instance, come from a related classification task with new fine-grained classes,
or it can have statistical distribution shift compared to the previously seen data. 
Each set of data that is presented to the model is referred to as a \textit{task}. 
Continual learning aims to create models which perform well on all seen tasks without the need to retrain from scratch when new data comes  \cite{Kirkpatrick_catastrophic_2017, Parisi_continual_2019}.
Continual learning 
has been demonstrated for a large breadth of tasks with real world data, including image and gesture classification \cite{DeLange_2022}
and wireless system design \cite{Sun_Learning_Continuously_2022},
and has gained increasing attention during recent years \cite{Kirkpatrick_catastrophic_2017, Parisi_continual_2019, Evron_catastrophic_2022, French_Catstrophic_1999, Doan_Forgetting_NTK_2021,bennani2020generalisation,DeLange_2022}.

Continual learning is related to learning under non-stationary distributions and adaptive filtering,
where iterative optimization methods and a stream of data are used to continuously adapt signal
models to unknown and possibly changing  environments \cite{sayed2008adaptive}.
Various phenomena of interest, 
such as financial time-series and target tracking,
often exhibit structural changes in signal characteristics over time. Hence, performance under non-stationary distributions has been  the focus in a number of scenarios; 
including randomly drifting unknowns in  distributed learning \cite{NosratiShamsiTaheriSedaaghi_2015},
switching system dynamics \cite{FoxSudderthJordanWillsky_2011,
DingShahrampirHealTarokh_2018,
KarimiButalZhaoKamalabadi_2022}.

The central issue in continual learning is \textit{forgetting}, 
which measures the performance degradation on previously learned tasks as new tasks are learnt by the model \cite{French_Catstrophic_1999, Evron_catastrophic_2022, Kirkpatrick_catastrophic_2017}.
If a model performs worse on the old tasks as new tasks are trained upon,
the model is said to exhibit \textit{catastrophic forgetting} \cite{French_Catstrophic_1999, Evron_catastrophic_2022, Kirkpatrick_catastrophic_2017}.
This emphasis on the performance on old tasks is what distinguishes continual learning apart from the long line of existing work focusing on estimation with non-stationary distributions,
where the estimator is expected to track the changing distribution, 
i.e., learn to perform well on the new distribution of data,
but not necessarily on data from the previous distributions.

Significant effort has been put into studying continual learning empirically
\cite{goodfellow2015empirical, nguyen2019understanding, farquhar2019robust, ramasesh2021anatomy}.
Compared to the breadth of existing empirical works,
theoretical analysis of continual learning lacks behind.
Nevertheless, 
bounds and explicit characterizations of continual learning performance have been recently presented for a range of models in order to close the gap between practice and theory,
e.g., focusing on the least-squares estimator \cite{lin_2023_theory}, 
the neural tangent regime 
\cite{bennani2020generalisation,Doan_Forgetting_NTK_2021},
and variants of stochastic gradient descent \cite{Evron_catastrophic_2022,lin_2023_theory,bennani2020generalisation}.
Here, we contribute to this line of  work by considering the continual learning problem from a distributed learning perspective.

In distributed learning,
optimization is performed over a network of computational nodes.
It not only supports learning over large scale models by spreading the computational load over multiple computational units, 
such as in edge computing \cite{wang_adaptive_2019},
but also provides an attractive framework for handling the emerging concerns for data security and privacy \cite{niknam_federated_2019, wang_privacy-preserving_2020}. 
Distributed learning is particularly attractive for scenarios where the data is already distributed over a network,
e.g., in sensor networks \cite{rabbat_distributed_2004, Kar_Distributed_2009,HuaNassifRicharWangSayed_2020,sayed2014adaptation},
or in dictionary learning where sub-dictionaries are naturally separated over the network
\cite{Sayed_Dictionary_distributed_2015}.

We focus on the successful distributed learning algorithm \cocoa{}~\cite{smith_cocoa_nodate}. 
\cocoa{} was developed from \cocoa{}-v1 \cite{jaggi2014communication}
and \cocoa{}\textsuperscript{+} \cite{ma2017distributed},
and has been extended into the fully decentralized algorithm \cola{} \cite{he_cola_2019}.
In \cocoa{}, nodes may utilize different local solvers with varying accuracies;  allowing exploration of different communication and computation trade-offs \cite{smith_cocoa_nodate}. 
Unlike the distributed learning frameworks  where data samples,
such as sensor readings, 
are distributed over the network 
\cite{sayed2014adaptation};
in \cocoa{},
the unknown model parameters to be estimated and the corresponding features
are distributed over the network, similar to \cite{tsitsiklis_distributed_1984,Khan2008Distributing}.

Our distributed continual learning framework with \cocoa{} is closely related to,
but nevertheless different from,
the typical framework for multitask learning over networks 
\cite{Skoglund_2021_RNN_Multitask, Sayed_2014_multitask_networks, Sayed_MTL_graphs_jsp_2020}. 
In particular, in our continual learning setting all nodes have the same task and this task changes over time for all nodes;  whereas in typical settings for multitask learning over networks
\cite{Skoglund_2021_RNN_Multitask, Sayed_2014_multitask_networks, Sayed_MTL_graphs_jsp_2020},
the individual nodes have different tasks which do not change over time. 
\balert{Successful examples of continual learning over networks have been presented \cite{yoon_federated_2021,dong_federated_2022,shenaj_asynchronous_2023}, under a range of constraints including asynchronous updates \cite{shenaj_asynchronous_2023} and privacy \cite{dong_federated_2022}. In contrast to these works focusing on empirical performance, we focus on providing analytical performance guarantees for \cocoa{}. }


%

We investigate how well 
\cocoa{} 
performs continual learning for a sequence of tasks where the data comes from a linear model for each task.
We focus on the \textit{generalization error} as our main performance metric.
The generalization error measures the output prediction error that the model makes on all the tasks,
using new data
unseen during training,
independent and identically distributed (i.i.d.) with the training data of the respective tasks.
See Section~\ref{sec:performanceMetrics} for a formal definition of the generalization error.
We present closed form expressions of the generalization error for a range of scenarios under isotropic Gaussian regressors.
\textbf{Our main contributions can be summarized \balert{as} follows:}
 
\begin{itemize}
    \item We provide exact analytical expressions for the generalization error of \cocoa{} under continual learning for the overparametrized case,
    as well as for the  scenario with a single update for each task.

    \item We show that \cocoa{} can perform continual learning through both analytical characterization and numerical illustrations.

    \item Our analytical results characterize the dependence of the generalization error on the network structure as well as the number of tasks and the task similarity. 
    
    \item We give sufficient conditions for a network structure to yield 
    zero generalization error and training error for a large number of tasks under stationary data distributions.

\end{itemize}

Our work provides analytical characterizations of the generalization error,
see, e.g.,
Theorem~\ref{thm:expected_gen_isoG}, Corollary~\ref{res:gen:equaldimensions},  Theorem~\ref{thm:similarity_generr}, 
Corollary~\ref{col:decomposition}
and Corollary~\ref{col:decomposition:modelps},
that have not been provided in the previous literature; and complement the  numerical studies of continual learning  with \cocoa{} in \cite{hellkvistOzcelikkaleAhlen_2022continualConf_arxiv}.
Our results extend the generalization error analysis in the single task setting of \cite{hellkvist_ozcelikkale_ahlen_linear_2021} to that of continual learning; and centralized continual learning setting of \cite{lin_2023_theory} to distributed learning. 
In contrast to \cite{Evron_catastrophic_2022} where the focus is on the training error
for the case where the unknown model parameter vector is the same for all tasks, 
i.e., only the regressors change over the tasks,
our analysis focuses on the generalization error when the tasks may possibly have different unknown model parameters.

Our closed form analytical expressions quantify the dependence of the generalization error to the 
number of samples per task, 
the number of nodes in the network,
and the number of unknowns governed by each node as well as similarities between tasks. 
Here similarities between tasks are measured through inner products between model parameter vectors, 
see Theorem~\ref{thm:expected_gen_isoG}.   
We analytically quantify how the task similarity affects whether the error increases or decreases as the number of tasks increase. 
These results reveal that the generalization error can be significantly reduced by adjusting the network size where the most favorable network size depends on task similarity and the number of tasks.
Furthermore, our numerical results illustrate that by reducing the number of \cocoa{} iterations for  each task, one 
may obtain lower generalization error than if \cocoa{} is run until convergence; further illustrating that   choosing \cocoa{} parameters that obtain the best continual learning performance is not straightforward.

    \begin{figure}[t]
        \centering
        \input{figures/sysfig}
        \caption{ 
            Distributed continual learning with \cocoa{},
            using $T_c$ iterations, 
            for a network of $K$ nodes
            operating on $T$ tasks.
            }
        \label{fig:system_fig}
    \end{figure}
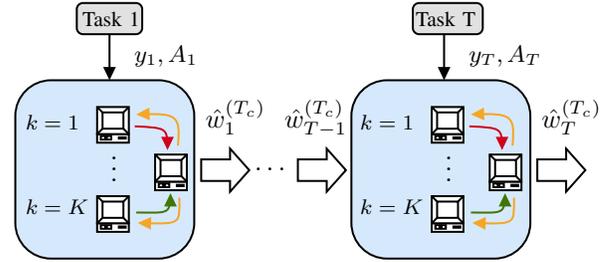

The rest of the paper is organized as follows:
Section~\ref{sec:problemStatement} presents the problem formulation.
Our characterization of the generalization error is provided in Section~\ref{sec:gen_err_analytical}
together with a range of example scenarios. 
We illustrate our findings with numerical results in Section~\ref{sec:numericalResults}. 
Finally, we conclude the paper in Section~\ref{sec:conclusions}.

\textbf{Notation:}
Let $\uvec, \qvec \inrbb{d\times 1}$ be real-valued vectors. 
We denote the Euclidean norm
as $\|\uvec\|=\sqrt{\langle \uvec, \uvec \rangle}$,
where $\langle \uvec, \qvec \rangle = \uvec\T \qvec$.
We denote the weighted norm and inner product with
$\|\uvec\|_{\Sigmamat}=\sqrt{\langle \uvec, \uvec \rangle_{\Sigmamat}}$,
where $\langle \uvec, \qvec \rangle_{\Sigmamat} = \uvec\T\Sigmamat\qvec$
for any  symmetric positive semi-definite matrix $\Sigmamat\inrbb{d\times d}$.
The Moore-Penrose pseudoinverse and transpose of a matrix $\Amat$ are denoted by $\Amat\p$ and $\Amat\T$, respectively.
The $p\times p$ identity matrix is denoted as $\eye{p}$.
%
%
%
The notation $\uvec\sim\Nc(\muvec, \Sigmamat)$
 denotes that $\uvec\inrbb{d\times 1}$ is a Gaussian random vector 
with the mean $\muvec\inrbb{d\times 1}$ and the covariance matrix $\Sigmamat\inrbb{d\times d}$.
For a vector of unknown parameters $\wvec^*  \inrbb{p\times 1}$, the superscript $^*$ is used to emphasize that this is the true value of the unknown vector.
For products where the lower limit exceeds the upper limit,
we use the convention $\prod_{\ell=\tau}^{\tau-1} \Mmat_{\ell} = \eye{p}$,
where $\Mmat_\ell\inrbb{p\times p}$ are non-zero matrices.

%% file: figures/sysfig.tex

\tikzset{every picture/.style={line width=0.75pt}} 

\begin{tikzpicture}[x=0.75pt,y=0.75pt,yscale=-0.75,xscale=0.75]

\draw  [fill={rgb, 255:red, 214; green, 233; blue, 252 }  ,fill opacity=1 ] (6,134) .. controls (6,120.75) and (16.75,110) .. (30,110) -- (102.33,110) .. controls (115.59,110) and (126.33,120.75) .. (126.33,134) -- (126.33,206) .. controls (126.33,219.25) and (115.59,230) .. (102.33,230) -- (30,230) .. controls (16.75,230) and (6,219.25) .. (6,206) -- cycle ;
\draw    (69.33,79) -- (69.33,107) ;
\draw [shift={(69.33,110)}, rotate = 270] [fill={rgb, 255:red, 0; green, 0; blue, 0 }  ][line width=0.08]  [draw opacity=0] (8.93,-4.29) -- (0,0) -- (8.93,4.29) -- cycle    ;
\draw [color={rgb, 255:red, 208; green, 2; blue, 27 }  ,draw opacity=1 ]   (86.33,141) .. controls (106.79,141.93) and (108.23,142) .. (108.33,154.07) ;
\draw [shift={(108.33,157)}, rotate = 270] [fill={rgb, 255:red, 208; green, 2; blue, 27 }  ,fill opacity=1 ][line width=0.08]  [draw opacity=0] (8.93,-4.29) -- (0,0) -- (8.93,4.29) -- cycle    ;
\draw [color={rgb, 255:red, 245; green, 166; blue, 35 }  ,draw opacity=1 ]   (90.46,132.89) .. controls (114.57,132.11) and (117.25,133.84) .. (115.33,154) ;
\draw [shift={(87.33,133)}, rotate = 357.88] [fill={rgb, 255:red, 245; green, 166; blue, 35 }  ,fill opacity=1 ][line width=0.08]  [draw opacity=0] (8.93,-4.29) -- (0,0) -- (8.93,4.29) -- cycle    ;
\draw [color={rgb, 255:red, 65; green, 117; blue, 5 }  ,draw opacity=1 ]   (87.33,199) .. controls (106.86,199) and (109.96,200.73) .. (110.3,188.91) ;
\draw [shift={(110.33,186)}, rotate = 90] [fill={rgb, 255:red, 65; green, 117; blue, 5 }  ,fill opacity=1 ][line width=0.08]  [draw opacity=0] (8.93,-4.29) -- (0,0) -- (8.93,4.29) -- cycle    ;
\draw [color={rgb, 255:red, 245; green, 166; blue, 35 }  ,draw opacity=1 ]   (91.37,206.1) .. controls (118.2,206.98) and (116.33,206.37) .. (116.33,189) ;
\draw [shift={(88.33,206)}, rotate = 1.91] [fill={rgb, 255:red, 245; green, 166; blue, 35 }  ,fill opacity=1 ][line width=0.08]  [draw opacity=0] (8.93,-4.29) -- (0,0) -- (8.93,4.29) -- cycle    ;
\draw  [fill={rgb, 255:red, 225; green, 225; blue, 225 }  ,fill opacity=1 ] (47.33,62.2) .. controls (47.33,59.88) and (49.21,58) .. (51.53,58) -- (87.13,58) .. controls (89.45,58) and (91.33,59.88) .. (91.33,62.2) -- (91.33,74.8) .. controls (91.33,77.12) and (89.45,79) .. (87.13,79) -- (51.53,79) .. controls (49.21,79) and (47.33,77.12) .. (47.33,74.8) -- cycle ;
\draw  [fill={rgb, 255:red, 255; green, 255; blue, 255 }  ,fill opacity=1 ] (61,146.22) -- (82.33,146.22) -- (82.33,152) -- (61,152) -- cycle ;
\draw  [fill={rgb, 255:red, 255; green, 255; blue, 255 }  ,fill opacity=1 ] (63.46,147.51) -- (66.47,147.51) -- (66.47,149.92) -- (63.46,149.92) -- cycle ;
\draw  [fill={rgb, 255:red, 255; green, 255; blue, 255 }  ,fill opacity=1 ] (75.61,148.38) -- (80.22,148.38) -- (80.22,149.22) -- (75.61,149.22) -- cycle ;
\draw  [fill={rgb, 255:red, 255; green, 255; blue, 255 }  ,fill opacity=1 ] (61,128) -- (82.33,128) -- (82.33,146.22) -- (61,146.22) -- cycle ; \draw   (64.64,131.64) -- (78.69,131.64) -- (78.69,142.58) -- (64.64,142.58) -- cycle ; \draw   (61,128) -- (64.64,131.64) ; \draw   (82.33,128) -- (78.69,131.64) ; \draw   (82.33,146.22) -- (78.69,142.58) ; \draw   (61,146.22) -- (64.64,142.58) ;
\draw  [fill={rgb, 255:red, 255; green, 255; blue, 255 }  ,fill opacity=1 ] (67.52,147.51) -- (70.53,147.51) -- (70.53,149.92) -- (67.52,149.92) -- cycle ;

\draw  [fill={rgb, 255:red, 255; green, 255; blue, 255 }  ,fill opacity=1 ] (61,206.22) -- (82.33,206.22) -- (82.33,212) -- (61,212) -- cycle ;
\draw  [fill={rgb, 255:red, 255; green, 255; blue, 255 }  ,fill opacity=1 ] (63.46,207.51) -- (66.47,207.51) -- (66.47,209.92) -- (63.46,209.92) -- cycle ;
\draw  [fill={rgb, 255:red, 255; green, 255; blue, 255 }  ,fill opacity=1 ] (75.61,208.38) -- (80.22,208.38) -- (80.22,209.22) -- (75.61,209.22) -- cycle ;
\draw  [fill={rgb, 255:red, 255; green, 255; blue, 255 }  ,fill opacity=1 ] (61,188) -- (82.33,188) -- (82.33,206.22) -- (61,206.22) -- cycle ; \draw   (64.64,191.64) -- (78.69,191.64) -- (78.69,202.58) -- (64.64,202.58) -- cycle ; \draw   (61,188) -- (64.64,191.64) ; \draw   (82.33,188) -- (78.69,191.64) ; \draw   (82.33,206.22) -- (78.69,202.58) ; \draw   (61,206.22) -- (64.64,202.58) ;
\draw  [fill={rgb, 255:red, 255; green, 255; blue, 255 }  ,fill opacity=1 ] (67.52,207.51) -- (70.53,207.51) -- (70.53,209.92) -- (67.52,209.92) -- cycle ;

\draw  [fill={rgb, 255:red, 255; green, 255; blue, 255 }  ,fill opacity=1 ] (100,178.22) -- (121.33,178.22) -- (121.33,184) -- (100,184) -- cycle ;
\draw  [fill={rgb, 255:red, 255; green, 255; blue, 255 }  ,fill opacity=1 ] (102.46,179.51) -- (105.47,179.51) -- (105.47,181.92) -- (102.46,181.92) -- cycle ;
\draw  [fill={rgb, 255:red, 255; green, 255; blue, 255 }  ,fill opacity=1 ] (114.61,180.38) -- (119.22,180.38) -- (119.22,181.22) -- (114.61,181.22) -- cycle ;
\draw  [fill={rgb, 255:red, 255; green, 255; blue, 255 }  ,fill opacity=1 ] (100,160) -- (121.33,160) -- (121.33,178.22) -- (100,178.22) -- cycle ; \draw   (103.64,163.64) -- (117.69,163.64) -- (117.69,174.58) -- (103.64,174.58) -- cycle ; \draw   (100,160) -- (103.64,163.64) ; \draw   (121.33,160) -- (117.69,163.64) ; \draw   (121.33,178.22) -- (117.69,174.58) ; \draw   (100,178.22) -- (103.64,174.58) ;
\draw  [fill={rgb, 255:red, 255; green, 255; blue, 255 }  ,fill opacity=1 ] (106.52,179.51) -- (109.53,179.51) -- (109.53,181.92) -- (106.52,181.92) -- cycle ;

\draw   (131,163.25) -- (150.4,163.25) -- (150.4,156) -- (163.33,170.5) -- (150.4,185) -- (150.4,177.75) -- (131,177.75) -- cycle ;
\draw   (196,163.25) -- (215.4,163.25) -- (215.4,156) -- (228.33,170.5) -- (215.4,185) -- (215.4,177.75) -- (196,177.75) -- cycle ;
\draw  [fill={rgb, 255:red, 214; green, 233; blue, 252 }  ,fill opacity=1 ] (232,134) .. controls (232,120.75) and (242.75,110) .. (256,110) -- (328.33,110) .. controls (341.59,110) and (352.33,120.75) .. (352.33,134) -- (352.33,206) .. controls (352.33,219.25) and (341.59,230) .. (328.33,230) -- (256,230) .. controls (242.75,230) and (232,219.25) .. (232,206) -- cycle ;
\draw    (294.33,79) -- (294.33,107) ;
\draw [shift={(294.33,110)}, rotate = 270] [fill={rgb, 255:red, 0; green, 0; blue, 0 }  ][line width=0.08]  [draw opacity=0] (8.93,-4.29) -- (0,0) -- (8.93,4.29) -- cycle    ;
\draw [color={rgb, 255:red, 208; green, 2; blue, 27 }  ,draw opacity=1 ]   (311.33,141) .. controls (331.79,141.93) and (333.23,142) .. (333.33,154.07) ;
\draw [shift={(333.33,157)}, rotate = 270] [fill={rgb, 255:red, 208; green, 2; blue, 27 }  ,fill opacity=1 ][line width=0.08]  [draw opacity=0] (8.93,-4.29) -- (0,0) -- (8.93,4.29) -- cycle    ;
\draw [color={rgb, 255:red, 245; green, 166; blue, 35 }  ,draw opacity=1 ]   (315.46,132.89) .. controls (339.57,132.11) and (342.25,133.84) .. (340.33,154) ;
\draw [shift={(312.33,133)}, rotate = 357.88] [fill={rgb, 255:red, 245; green, 166; blue, 35 }  ,fill opacity=1 ][line width=0.08]  [draw opacity=0] (8.93,-4.29) -- (0,0) -- (8.93,4.29) -- cycle    ;
\draw [color={rgb, 255:red, 65; green, 117; blue, 5 }  ,draw opacity=1 ]   (312.33,199) .. controls (331.86,199) and (334.96,200.73) .. (335.3,188.91) ;
\draw [shift={(335.33,186)}, rotate = 90] [fill={rgb, 255:red, 65; green, 117; blue, 5 }  ,fill opacity=1 ][line width=0.08]  [draw opacity=0] (8.93,-4.29) -- (0,0) -- (8.93,4.29) -- cycle    ;
\draw [color={rgb, 255:red, 245; green, 166; blue, 35 }  ,draw opacity=1 ]   (316.37,206.1) .. controls (343.2,206.98) and (341.33,206.37) .. (341.33,189) ;
\draw [shift={(313.33,206)}, rotate = 1.91] [fill={rgb, 255:red, 245; green, 166; blue, 35 }  ,fill opacity=1 ][line width=0.08]  [draw opacity=0] (8.93,-4.29) -- (0,0) -- (8.93,4.29) -- cycle    ;
\draw  [fill={rgb, 255:red, 255; green, 255; blue, 255 }  ,fill opacity=1 ] (286,146.22) -- (307.33,146.22) -- (307.33,152) -- (286,152) -- cycle ;
\draw  [fill={rgb, 255:red, 255; green, 255; blue, 255 }  ,fill opacity=1 ] (288.46,147.51) -- (291.47,147.51) -- (291.47,149.92) -- (288.46,149.92) -- cycle ;
\draw  [fill={rgb, 255:red, 255; green, 255; blue, 255 }  ,fill opacity=1 ] (300.61,148.38) -- (305.22,148.38) -- (305.22,149.22) -- (300.61,149.22) -- cycle ;
\draw  [fill={rgb, 255:red, 255; green, 255; blue, 255 }  ,fill opacity=1 ] (286,128) -- (307.33,128) -- (307.33,146.22) -- (286,146.22) -- cycle ; \draw   (289.64,131.64) -- (303.69,131.64) -- (303.69,142.58) -- (289.64,142.58) -- cycle ; \draw   (286,128) -- (289.64,131.64) ; \draw   (307.33,128) -- (303.69,131.64) ; \draw   (307.33,146.22) -- (303.69,142.58) ; \draw   (286,146.22) -- (289.64,142.58) ;
\draw  [fill={rgb, 255:red, 255; green, 255; blue, 255 }  ,fill opacity=1 ] (292.52,147.51) -- (295.53,147.51) -- (295.53,149.92) -- (292.52,149.92) -- cycle ;

\draw  [fill={rgb, 255:red, 255; green, 255; blue, 255 }  ,fill opacity=1 ] (286,206.22) -- (307.33,206.22) -- (307.33,212) -- (286,212) -- cycle ;
\draw  [fill={rgb, 255:red, 255; green, 255; blue, 255 }  ,fill opacity=1 ] (288.46,207.51) -- (291.47,207.51) -- (291.47,209.92) -- (288.46,209.92) -- cycle ;
\draw  [fill={rgb, 255:red, 255; green, 255; blue, 255 }  ,fill opacity=1 ] (300.61,208.38) -- (305.22,208.38) -- (305.22,209.22) -- (300.61,209.22) -- cycle ;
\draw  [fill={rgb, 255:red, 255; green, 255; blue, 255 }  ,fill opacity=1 ] (286,188) -- (307.33,188) -- (307.33,206.22) -- (286,206.22) -- cycle ; \draw   (289.64,191.64) -- (303.69,191.64) -- (303.69,202.58) -- (289.64,202.58) -- cycle ; \draw   (286,188) -- (289.64,191.64) ; \draw   (307.33,188) -- (303.69,191.64) ; \draw   (307.33,206.22) -- (303.69,202.58) ; \draw   (286,206.22) -- (289.64,202.58) ;
\draw  [fill={rgb, 255:red, 255; green, 255; blue, 255 }  ,fill opacity=1 ] (292.52,207.51) -- (295.53,207.51) -- (295.53,209.92) -- (292.52,209.92) -- cycle ;

\draw  [fill={rgb, 255:red, 255; green, 255; blue, 255 }  ,fill opacity=1 ] (325,178.22) -- (346.33,178.22) -- (346.33,184) -- (325,184) -- cycle ;
\draw  [fill={rgb, 255:red, 255; green, 255; blue, 255 }  ,fill opacity=1 ] (327.46,179.51) -- (330.47,179.51) -- (330.47,181.92) -- (327.46,181.92) -- cycle ;
\draw  [fill={rgb, 255:red, 255; green, 255; blue, 255 }  ,fill opacity=1 ] (339.61,180.38) -- (344.22,180.38) -- (344.22,181.22) -- (339.61,181.22) -- cycle ;
\draw  [fill={rgb, 255:red, 255; green, 255; blue, 255 }  ,fill opacity=1 ] (325,160) -- (346.33,160) -- (346.33,178.22) -- (325,178.22) -- cycle ; \draw   (328.64,163.64) -- (342.69,163.64) -- (342.69,174.58) -- (328.64,174.58) -- cycle ; \draw   (325,160) -- (328.64,163.64) ; \draw   (346.33,160) -- (342.69,163.64) ; \draw   (346.33,178.22) -- (342.69,174.58) ; \draw   (325,178.22) -- (328.64,174.58) ;
\draw  [fill={rgb, 255:red, 255; green, 255; blue, 255 }  ,fill opacity=1 ] (331.52,179.51) -- (334.53,179.51) -- (334.53,181.92) -- (331.52,181.92) -- cycle ;

\draw   (357,163.25) -- (376.4,163.25) -- (376.4,156) -- (389.33,170.5) -- (376.4,185) -- (376.4,177.75) -- (357,177.75) -- cycle ;
\draw  [fill={rgb, 255:red, 225; green, 225; blue, 225 }  ,fill opacity=1 ] (273.33,62.2) .. controls (273.33,59.88) and (275.21,58) .. (277.53,58) -- (316.13,58) .. controls (318.45,58) and (320.33,59.88) .. (320.33,62.2) -- (320.33,74.8) .. controls (320.33,77.12) and (318.45,79) .. (316.13,79) -- (277.53,79) .. controls (275.21,79) and (273.33,77.12) .. (273.33,74.8) -- cycle ;

\draw (132,121) node [anchor=north west][inner sep=0.75pt]    {$\hat{w}_{1}^{( T_{c})}$};
\draw (165,167) node [anchor=north west][inner sep=0.75pt]    {$\dotsc $};
\draw (185,121) node [anchor=north west][inner sep=0.75pt]    {$\hat{w}_{T-1}^{( T_{c})}$};
\draw (358,121) node [anchor=north west][inner sep=0.75pt]    {$\hat{w}_{T}^{( T_{c})}$};
\draw (11,131) node [anchor=north west][inner sep=0.75pt]  [font=\footnotesize]  {$k=1$};
\draw (11,188) node [anchor=north west][inner sep=0.75pt]  [font=\footnotesize]  {$k=K$};
\draw (70.51,68.89) node  [font=\footnotesize] [align=left] {Task 1};
\draw (84,85) node [anchor=north west][inner sep=0.75pt]  [font=\small]  {$y_{1} ,A_{1}$};
\draw (236,131) node [anchor=north west][inner sep=0.75pt]  [font=\footnotesize]  {$k=1$};
\draw (236,188) node [anchor=north west][inner sep=0.75pt]  [font=\footnotesize]  {$k=K$};
\draw (295.51,68.89) node  [font=\footnotesize] [align=left] {Task T};
\draw (309,85) node [anchor=north west][inner sep=0.75pt]  [font=\small]  {$y_{T} ,A_{T}$};
\draw (68,149) node [anchor=north west][inner sep=0.75pt]    {$\vdots $};
\draw (293,149) node [anchor=north west][inner sep=0.75pt]    {$\vdots $};

\end{tikzpicture}

%% file: problem_statement.tex
\section{Problem Statement}\label{sec:problemStatement}

\subsection{Definition of a Task}
We consider the continual learning problem of solving $T$ tasks. 
The {\emph{training data}} for task $t$, $1 \leq t\leq  T$ is denoted by $(\yvec_t, \Amat_t)$ and consists of the regressor matrix 
$\Amat_t\inrbb{n_t\times p}$   and the corresponding vector of outputs
$\yvec_t\inrbb{n_t\times 1}$. 
The observations $\yvec_t $ come from the following noisy linear  model
\begin{equation}\label{eqn:model_matrix}
    \yvec_t = \Amat_t\wvec_t^* + \zvec_t.
\end{equation}
where each task has an unknown model parameter vector 
$\wvec_t^*\inrbb{p\times 1}$, 
and $\zvec_t\inrbb{n_t\times 1}$ denotes the noise.
The noise vectors
$\zvec_t$'s are independently distributed
with $\zvec_t\sim\Nc(\bm 0, \sigma_t^2 \eye{n_t})$,
and they are statistically independent from the regressors.
We focus on the following regressor model: 

\begin{assumption}\label{assume:standard_Gaussian_regressors}
    The matrices $\Amat_t$, $t=1,\,\dots,\, T$,
    are statistically independent and 
    have independently and identically distributed (i.i.d.) standard Gaussian entries. In particular, if $\avec\T\inrbb{1\times p}$ is a row of $\Amat_t$,
then $\avec \sim \Nc(\bm 0, \eye{p})$.

\end{assumption}

\kern0.25em
\noindent 

Task $t$ corresponds to finding  an estimate $\what$ so that $\yvec_t \approx \Amat_t \what$. In particular, for  task $t$, the aim is to minimize the training error 
 given by the mean-squared-error (MSE) 
\begin{equation}\label{eqn:training_MSE}
    \Ltrain{t}(\what) 
    = \frac{1}{2 n_t}\left\|\Amat_t \what - \yvec_t\right\|^2,
\end{equation}
where $\|\cdot\|$ denotes the Euclidean norm. 
Note that  the scaling of $\frac{1}{2}$  is included here
for  notational convenience later.

\subsection{Continual Learning Problem}\label{sec:prob:contLearningIntro}
We study the continual learning setting, 
where  training data for the tasks arrive sequentially, and   the data  for only one task is available at a given time.
In particular, 
at a given time instant $t$,
we only have access to $(\yvec_t, \Amat_t)$ for a particular task $t$ but not the data for the other tasks. The goal is to find  an estimate $\what$ that performs well over all tasks  $1 \leq t\leq  T$. 
\galert{%
Hence,  we would like to find a single  $\what$ such that 
\begin{align} \label{eqn:system:alltasks}
    \Amat_t \what \approx \yvec_t, \, \forall\, t.  
\end{align}
We note that if all the data were available at once,
instead of arriving sequentially, one may find a solution for \eqref{eqn:system:alltasks} by solving the minimization problem 
$
    \min_{\wvec} 
    \frac{1}{2 N}
    \sum_{t=1}^T 
    \| \Amat_t \wvec - \yvec_t \|^2,
$
where $N = \sum_{t=1}^T n_t$. 
Hence, the aim of the continual learning setting can be seen as to
perform as close as possible to this solution,
despite the sequential nature of data arrival. 
Further details on this benchmark can be found in Section~\ref{sec:offlineCentralized}.}

We focus on the scenarios where the task identity,
i.e., $t$, is not available and it does not need to be inferred;
hence we look for a single $\what$ for all tasks. 
This type of continual setting is referred as domain-incremental learning;
and it is suited to the scenarios where  the structure of the tasks is  the same but the input distribution changes \cite{VenTolias_2019}.

Precise definitions of the performance metrics for continual learning are presented in Section~\ref{sec:performanceMetrics}. We provide a detailed description  of the distributed iterative algorithm \cocoa{} employed to find the estimate in Section~\ref{sec:distributed_learning}, see also Fig.~\ref{fig:system_fig}.
An overview of some of the important variables of the problem formulation is provided in Table~\ref{table:variables}.

\subsection{Performance Metrics}\label{sec:performanceMetrics}

Let $\what_\tau$ denote an estimate obtained by training on
a dataset $\traindata{\tau}=\{\yvec_t,\, \Amat_t\}_{t=1}^\tau$ 
where $1\leq \tau \leq T$.
Note that the subscript $\tau$ of the estimate
$\what_\tau$ indicates that the data up to and 
including task $\tau$ has been used to create the estimate. 

To characterize how well an estimate achieves continual learning, we consider forgetting from two different perspectives: 
\textit{training error}
and
\textit{generalization error}.

\subsubsection{Training Error}
Forgetting in terms of 
training error
measures the output error that an estimate makes on the training data of the tasks previously seen. 
It is defined as \cite{Evron_catastrophic_2022}
\begin{equation}\label{eqn:forgetting_def}
    \Forgetting{\tau}(\what_\tau) 
    = \frac{1}{\tau} \sum_{t=1}^\tau \Ltrain{t}
        \left(\what_\tau\right)
    = \frac{1}{\tau} \sum_{t=1}^\tau \frac{1}{2 n_t}
        \left\|\Amat_t \what_\tau - \yvec_t \right\|^2.
\end{equation}
Hence, 
$ \Forgetting{\tau}(\what_\tau) $ 
is the average MSE over
the training data used to create the estimate $\what_\tau$,
i.e., from the tasks $t=1,\,\dots,\,\tau$.

We define the \textit{expected training error}
of $\what_\tau$ over the distribution of the training data $\traindata{\tau}$  
as 
\begin{align}
    F_\tau(\what_\tau)
    & = \eunder{\traindata{\tau}}\left[
        f_\tau \left( \what_\tau \right)
    \right] \\    
    & = \frac{1}{\tau} 
        \sum_{t=1}^\tau
        \frac{1}{2 n_t}
        \eunder{\traindata{\tau}}\left[
            \left\|\Amat_t \what_\tau - \yvec_t \right\|^2
        \right].\label{eqn:expected_forgetting}
\end{align}
Since $\what_\tau$ depends on data up to and including task $\tau$, the expectation in \eqref{eqn:expected_forgetting} is over $\traindata{\tau}=\{\yvec_t,\, \Amat_t\}_{t=1}^\tau$
and not merely over only the latest task 
$(\yvec_\tau,\, \Amat_\tau)$.

\begin{table}
\caption{Overview of Important Variables.}
\label{table:variables}
\centering
\begin{tabular}{ |m{0.12\linewidth}|m{0.76\linewidth}| } 
    \hline $\wvec_t^*$ 
        & true value of the unknown parameter for task t \\
    \hline $(\yvec_t, \Amat_t)$ 
        & training data for task $t$ \\ 
     \hline $\traindata{t}$ 
        & $\{\yvec_{t'},\, \Amat_{t'}\}_{t'=1}^t$ \\
    \hline $\what_t$
        & parameter estimate found using $\traindata{t}$ \\
    \hline $\what_t^{(i)}$
        & parameter estimate found by \cocoa{} using $\traindata{t}$ and $i$ iterations for task $t$ \\ 
    \hline
       $\what_{t,[k]}^{(i)}$
        & partition of $\what_t^{(i)}$ at node k of the network \\ 
    \hline
\end{tabular}
\end{table}

\subsubsection{Generalization Error}
The generalization error
measures the performance on data unseen during training.
Let $y_{t,\text{new}} = \avec_{t,\text{new}}\T \wvec_t^* + z_{t,\text{new}}$,
be a new unseen sample for task $t$,
i.i.d. with the samples of training data of that task,
i.e., $\avec_{t,\text{new}}\sim\Nc(\bm 0, \eye p)$
and $z_{t,\text{new}}\sim\Nc(0, \sigma_t^2)$.
Consider the estimate $\what_\tau$. 
The generalization error associated with $\what_\tau$ over tasks $\trange$,
is defined as
\begin{align}
    & \Generalization{T}(\what_\tau) 
    = \frac{1}{T} \sum_{t=1}^T \eunder{y_{t,\text{new}}, \avec_{t,\text{new}}}
        \left[ \left(\avec_{t,\text{new}}\T \what_\tau - y_{t,\text{new}}\right)^2 \right] \\
    & \quad = \frac{1}{T} \sum_{t=1}^T \eunder{\avec_{t,\text{new}}, z_{t,\text{new}}}
        \left[ \left(\avec_{t,\text{new}}\T \left(\what_\tau - \wvec_t^*\right) - z_{t,\text{new}}\right)^2 \right] \\
    \begin{split}
    & \quad = \frac{1}{T} \sum_{t=1}^T
        \left(\what_\tau - \wvec_t^* \right)\T\!\!
        \eunder{\avec_{t,\text{new}}} \left[
            \avec_{t,\text{new}} \avec_{t,\text{new}}\T \right]
        \left(\what_\tau - \wvec_t^* \right) \\
        & \qquad\qquad + \eunder{z_{t,\text{new}}} \left[z_{t,\text{new}}^2 \right]
    \end{split}\\
    & \quad = \frac{1}{T} \sum_{t=1}^T 
        \left\|\what_\tau - \wvec_t^*\right\|^2 
        + \sigma_t^2. \label{eqn:gen_err_def}
\end{align}
Forgetting is typically defined only on the tasks whose data is used for constructing the estimate \cite{Evron_catastrophic_2022, Kirkpatrick_catastrophic_2017},
see for instance the summation in \eqref{eqn:forgetting_def} which runs up to $\tau$. 
On the other hand, here we have chosen to define the generalization error in \eqref{eqn:gen_err_def} as the  performance over all tasks $1 \leq t \leq T$ instead of only up to the last seen task $\tau$. 
In addition to measuring the forgetting in terms of generalization on the tasks seen,
this definition allows us to keep track of how well the estimate performs for unseen tasks and leads to interesting conclusions in terms of effect of task similarity, see  Remark~\ref{remark:generalization:unseenTasks}.

We define the \textit{expected generalization error} over the distribution of the training data $\traindata{\tau}$ as
\begin{align}
    G_T(\what_\tau) 
    & = \eunder{\traindata{\tau}}
        \left[\Generalization{T}(\what_\tau)\right] \\
    & = \frac{1}{T}\sum_{t=1}^T
        \eunder{\traindata{\tau}}\left[
            \left\| \what_\tau - \wvec_t^* \right\|^2
        \right]
        + \sigma_t^2. \label{eqn:expected_gen_err_def}
\end{align}
Our main results in Theorem~\ref{thm:expected_gen_isoG} provide an analytical characterization of $G_T(\cdot)$  when the estimate 
$\what_\tau$ is obtained using the distributed algorithm \cocoa{}. 

\begin{algorithm}[t]
    \mycocoa
\end{algorithm}

\subsection{Distributed Continual Learning with \cocoa{}}
\label{sec:distributed_learning}

\subsubsection{Overview}
Consider  task $t$ with data $(\yvec_t, \Amat_t)$.
We minimize the MSE in \eqref{eqn:training_MSE}
in a distributed and iterative fashion using the algorithm \cocoa{}~\cite{smith_cocoa_nodate},
see  Figure~\ref{fig:system_fig} and Algorithm~\ref{alg:continual_cocoa}. 
For task $t$, the regressor matrix $\Amat_t$ and the initial estimate for the unknown vector is distributed over $K$ nodes.  Then, $T_c$  iterations of \cocoa{} are used to update the parameter estimate iteratively in order to minimize \eqref{eqn:training_MSE}. When a new task comes, the old parameter estimate is used as the initialization and the procedure is repeated.  
The below presentation of \cocoa{} includes references to the line numbers in Algorithm~\ref{alg:continual_cocoa}.

\subsubsection{Distribution over nodes}
We now describe how the data is distributed over the network. 
The regressor matrix $\Amat_t$ is distributed over a network of $K$ nodes
by column-wise partitions
such that each node governs an exclusive set of columns of $\Amat_t$.
The partitioning of $\Amat_t$ is given by
\begin{equation}\label{eqn:partitioning_A}
    \Amat_t
    =  \begin{bmatrix} \Amat_{t,[1]} & \cdots & \Amat_{t,[K]} \end{bmatrix}.
\end{equation}
We have $\Amat_{t,[k]}\inrbb{n_t\times p_k}$,  $\forall t$, hence  the number of columns in the $k$\textsuperscript{th} submatrix of $\Amat_t$
is  $p_k\in \Nbb$, $\forall t$.

Let us denote the model parameter estimate obtained after training on the data of task $t$ with $i$  iterations of \cocoa{}  by $\what_t^{(i)}\inrbb{p\times 1}$.
The column-wise partitioning of $\Amat_t$ corresponds to a row-wise partitioning in  $\what_t^{(i)}$ over the nodes as follows
\begin{equation}\label{eqn:wvec_k_stacked}
    \what_t^{(i)} 
    = \begin{bmatrix} \what_{t,[1]}^{(i)} \\ \vdots \\ \what_{t,[K]}^{(i)}\end{bmatrix},
\end{equation}
where $\what_{t,[k]}^{(i)}\inrbb{p_k\times 1}$ is the unknown model parameter partition at node $k$.

\subsubsection{Learning over tasks}
Let us have task $t-1$.  We run \cocoa{} for $T_c$ iterations to obtain the estimate $\what_{t-1}^{(T_c)}$.
When data for task $t$ comes,
i.e., $(\yvec_t, \Amat_t)$,
the algorithm is initialized with the estimate learnt with the data of the previous task,
i.e.,
\begin{equation}
    \what_t^{(0)} = \what_{t-1}^{(T_c)},
\end{equation}
see line 2.
For the first task, i.e. $t=1$,
we use $\what_1^{(0)} = \what_0^{(T_c)} = \bm 0$ as initialization.
In \cocoa{}, an auxiliary variable $ \vvec_{t}$ keeps track of  the contribution of each node to the current estimate of the observations $\yvec_t$.
Partitions of  $ \vvec_{t}$ are initialized as 
\begin{equation}
 \vvec_{t,[k]}^{(0)} = K \Amat_{t,[k]} \what_{t,[k]}^{(0)},\quad \forall k, 
 \end{equation}
see line 3. 
After $T_c$ iterations of \cocoa{} for task $t$,
 $\what_t^{(T_c)}$ is outputted 
as the estimate of the parameter vector,
see line 10.

\subsubsection{\cocoa{} iterations}\label{sec:problemFormulation:cocoa:iterations}
For each iteration $(i)$,
a central unit shares the network's aggregated current estimate of the observations $\yvec_t$,
denoted by $\vbar_t^{(i)}\inrbb{n_t\times 1}$ (line 5).
Each node then computes its local updates 
    $\Delta \what_{t,[k]}^{(i)}\inrbb{p_k\times 1}$ (line 7),  updates its  estimate of the parameter partition
    $\what^{(i)}_{t,[k]}$ (line 8) and the local estimate of $\yvec$, i.e., $\vvec_{t,[k]}^{(i)}\inrbb{n_t\times 1}$ (line 9).

In order to  minimize $\eqref{eqn:training_MSE}$, the node $k$'s update in line 7
is computed by solving the following local subproblem \cite{smith_cocoa_nodate}
\begin{align}\begin{split}
    \min_{\Delta \what_{t,[k]}^{(i)}} 
        \frac{1}{2K n_t}\left\|\vbar_t^{(i)} - \yvec_t\right\|^2
        + \frac{\subproblemp}{2 n_t} \left\|\Amat_{t,[k]}
            \Delta \what_{t,[k]}^{(i)} \right\|^2
        \\
        + \frac{1}{n_t}\left( \vbar_{t}^{(i)} - \yvec_t \right)\T \Amat_{t,[k]} \Delta \what_{t,[k]}^{(i)}.
\end{split}
\end{align}
This is a convex problem in $\Delta \what_{t,[k]}^{(i)}$.
Setting its derivative with respect to $\Delta \what_{t,[k]}^{(i)}$ to zero,
one arrives at the following
\begin{equation}\label{eqn:cocoa:normalEquations}
    \sigma' \Amat_{t,[k]}\T \Amat_{t,[k]} \Delta\what_{t,[k]}^{(i)} 
    = \Amat_{t,[k]}\T \left(\yvec_t - \vbar_t^{(i)}\right).
\end{equation}
The expression in line 7 is  the minimum $\ell_2$-norm solution to \eqref{eqn:cocoa:normalEquations}
where $(\cdot)\p$ in line 7  denotes the Moore-Penrose pseudoinverse.

In \cite{smith_cocoa_nodate},
\cocoa{} is presented with the hyperparameters $\subproblemp$ and $\aggregationp$,
referred to as the subproblem and aggregation parameters,
respectively.
We have set 
$\subproblemp=\aggregationp K$,
and let $\aggregationp \in (0, 1]$,
as these are considered safe choices \cite{he_cola_2019}.
As a result,
the specific value of $\aggregationp$ washes out to give the explicit expressions in Algorithm~\ref{alg:continual_cocoa}.

\subsection{A benchmark: Offline and centralized solution}\label{sec:offlineCentralized}
We compare the performance of the \cocoa{} solution with the following offline and centralized version of the problem. 
In particular, consider the setting where one has access to all the tasks' training data at once,
hence one minimizes MSE over all that data simultaneously,
i.e.,
$
    \min_{\wvec} 
    \frac{1}{2 N}
    \sum_{t=1}^T 
    \left\| \Amat_t \wvec - \yvec_t\right \|^2,
$
where $N = \sum_{t=1}^T n_t$.
The minimum $\ell_2$-norm solution, i.e., the standard least-squares solution, is then given by,
\begin{equation}\label{eqn:offline_centralized_solution}
    \offlinesoln = 
    \begin{bmatrix}
        \Amat_1 \\
        \vdots\\
        \Amat_T
    \end{bmatrix}\p
    \begin{bmatrix}
        \yvec_1\\\vdots\\\yvec_T.
    \end{bmatrix}.
\end{equation}
We refer to $\offlinesoln$
as the 
\textit{offline and centralized solution}, and we use it to compare the performance of \cocoa{} with.

%% file: main_results.tex
\section{Generalization Error }\label{sec:gen_err_analytical}
In this section, we present  closed form expressions for the generalization error of the estimates produced by Algorithm~\ref{alg:continual_cocoa} in the continual learning setting for a range of scenarios.
\subsection{Preliminaries}\label{sec:gen_err_analytical:prel}
We begin by presenting Lemma~\ref{lemma:recursion},
which describes the estimate $\what_t^{(1)}$,
i.e., the estimate found by Algorithm~\ref{alg:continual_cocoa}
after the first iteration when 
initialized with the parameter estimate from the previous task, 
i.e., $\what_{t-1}^{(T_c)}$. 
Note that Lemma~\ref{lemma:recursion} and also some other intermediate results, such as  Lemma~\ref{lemma:recursion_sum_of_alphas}, do not rely on Assumption~\ref{assume:standard_Gaussian_regressors}.  Hence, for the sake of clarity,  we refer to Assumption~\ref{assume:standard_Gaussian_regressors} in our results explicitly whenever it is needed.

\begin{lemma}\label{lemma:recursion}
    The solution $\what_t^{(1)}$
    after first iteration of Algorithm~\ref{alg:continual_cocoa}
    is given by the following expression,
    \begin{equation}\label{eqn:recursion}
        \what_t^{(1)} = \Pmat_t \what_{t-1}^{(T_c)} + \Abar_t \yvec_t,
    \end{equation}
    where
    \begin{equation}\label{eqn:P:definition}
        \Pmat_t = \eye{p} - \Abar_t \Amat_t \inrbb{p\times p}, ~
     \end{equation}
     \begin{equation}\label{eqn:Abar:definition}
        \Abar_t 
        = \frac{1}{K}
            \begin{bmatrix}
                \Amat_{t,[1]}\p \\ 
                \vdots\\
                \Amat_{t,[K]}\p
            \end{bmatrix}\inrbb{p\times n_t}.
    \end{equation}
    Furthermore,
    for any $\uvec\inrbb{p\times 1}$,
    \begin{equation}\label{eqn:recursion_minus_wi}
        \what_t^{(1)} \! - \uvec
        = \!\Pmat_t\!\left(\what_{t-1}^{(T_c)} \! - \! \uvec\right)
        + \Abar_t\Amat_t\!\left(\wvec_t^* \! - \! \uvec\right)
        + \Abar_t \zvec_t.
    \end{equation}
    
\end{lemma}

Proof: This result has been partially presented in \cite[Sec.~3]{hellkvistOzcelikkaleAhlen_2022continualConf_arxiv}, \cite[Sec.~IV]{hellkvist_ozcelikkale_ahlen_linear_2021}.
Due to slightly different assumptions therein, we present the proof in Appendix~\ref{proof:lemma:recursion} for the sake of completeness.

The recursion in \eqref{eqn:recursion_minus_wi} is used to present closed form expressions of the generalization error in Theorem~\ref{thm:expected_gen_isoG},
and asymptotic analyses of the generalization and training error
in Section~\ref{sec:task-similarity}.
%
The following result gives a sufficient condition
for which the algorithm converges in the first iteration.

\begin{lemma}\label{res:one-step-convergence}
    If the partitions $\Amat_{t,[k]}\inrbb{n_t\times p_k}$
    have full row rank,
    i.e., $\rank(\Amat_{t,[k]}) = n_t$,
    then Algorithm~\ref{alg:continual_cocoa}
    converges in the first iteration
    to the solution given by \eqref{eqn:recursion}
    i.e.,
    \begin{equation}\label{eqn:T_c_equals_T_c_1}
        \what_t^{(T_c)} = \what_t^{(1)},~ T_c \geq 1.
    \end{equation}

\end{lemma}

Proof: This result is proven in \cite[Lemma~1]{hellkvistOzcelikkaleAhlen_2022continualConf_arxiv}. 
Note that the first half of \cite[Lemma~1]{hellkvistOzcelikkaleAhlen_2022continualConf_arxiv},
i.e., the result here, does not require the stationarity condition \cite[Assumption~1]{hellkvistOzcelikkaleAhlen_2022continualConf_arxiv}. 

We use 
one
of the following assumptions in most of our analytical development: 
\begin{assumption}\label{assume:broad_and_large_Tc}
  $p_k > n_t+1,~ \forall t, k$.
\end{assumption}
\begin{assumption}\label{assume:T_c_is_one}
    $T_c = 1$.
\end{assumption}

\noindent We explicitly refer to these assumptions in our results as appropriate.

Recall that for a given task $t$, all nodes have the same number of data samples $n_t$ but possibly different number of model parameters. Assumption~\ref{assume:broad_and_large_Tc} corresponds to the case of  overparameterized local models, i.e., at each node $k$ there is a large number of model parameters $p_k$ compared to the number of data samples $n_t$. 
\balert{Many modern learning models operate in the overparameterized regime \cite{zhang_understanding_2017},
and massively overparametrized models have been very successful, see for instance \cite[Table 1]{zhang_understanding_2017}.
A number of works have focused on the overparametrized setting under distributed learning, investigating  
 theoretical guarantees for communication efficiency \cite{song_distributed_nodate,khanduri_decentralized_2022,zhang_distributed_2019}, convergence  \cite{qin_decentralized_2022, deng_local_nodate} and generalization error\cite{hellkvist_ozcelikkale_ahlen_linear_2021}.}
In addition to the empirical success of large overparametrized models \cite[Table 1]{zhang_understanding_2017}, analytical characterization of the generalization error of overparametrized linear models has also been the focus of many recent works, including the centralized scenario, e.g. \cite{nakkiran2020optimal,hastie2020surprises, Sahai_Harmless}, and distributed scenario without continual learning \cite{hellkvist_ozcelikkale_ahlen_linear_2021}. 
In this article, we contribute to these lines of work by focusing on a distributed and continual learning setting.

Assumption~\ref{assume:T_c_is_one} corresponds to the setting that only one update to  the parameter estimate is performed for a given task,
\balert{i.e., there is one round of communication between the nodes.} 
\balert{%
Our work here can be interpreted as the generalization of the centralized continual learning scenario in \cite{lin_2023_theory},  which also performs a one step update,  to the distributed case.
See Section~\ref{sec:comparison:centralized} for further discussions.}
\balert{In distributed learning, the scenarios with one round of communication,  is referred to as the one-shot setting \cite{stich_local_2019, salehkaleybar_one-shot_nodate, sharifnassab_order_nodate}.}
\balert{%
In practice, this type of setting may be used when the cost of communication is too high;
or when the computational memory load and the related energy constraints make it difficult for the nodes to perform multiple updates;
or when the data comes in real-time and it cannot be stored for the subsequent updates due to memory constraints.
Distributed one-shot settings have been the main focus of a number of works, such as \cite{heinze_loco_2015, heinze_dualloco_2016,salehkaleybar_one-shot_nodate, sharifnassab_order_nodate}.  
For some scenarios, one-shot solutions
have been shown to outperform or match standard solutions,
including the mixture weight method,
which uses one-shot communication and less resources overall while achieving a performance on the level of the standard gradient-based approach with multiple rounds of communication \cite{mcdonald_efficient_2009};
approaches based on averaging,
which use one-shot communication but achieve the best possible error rate decay achievable by a centralized algorithm \cite{zhang_comunication-efficient_2013};
and one-shot averaging with stochastic gradient, which can achieve the optimal asymptotic convergence rate \cite{spiridonoff_communication-efficient_2021}. }
\balert{In a similar manner,
running \cocoa{} with $T_c=1$ can 
in some scenarios 
provide better generalization performance compared to having a relatively high $T_c$, see the discussion for high number of tasks in  Section~\ref{sec:num:Tc}.}

\begin{remark}\label{rem:iso-G-large_Tc}
    If Assumption~\ref{assume:standard_Gaussian_regressors} 
    and Assumption \ref{assume:broad_and_large_Tc} hold,
    then the partitions $\Amat_{t,[k]}$ are full row-rank with probability 1,
    hence \eqref{eqn:T_c_equals_T_c_1} holds with probability 1. As a result, 
    the expected generalization error $G_T(\what_t^{(T_c)})$ 
    given by Algorithm~\ref{alg:continual_cocoa} with $T_c=1$
    is the same for all $T_c \geq 1$.
\end{remark}

    The results presented in this paper are often functions of the problem dimensions.
    In particular,
    the dimensions of the local partitions $\Amat_{t,[k]}\inrbb{n_t \times p_k}$,
    i.e., the number of samples per task $n_t$,
    and the number of model parameters per node $p_k$.
    Thus we introduce the following notation for the nodes $\krange$,
    and the tasks $\trange$,
    \begin{align}
        r_{t,k} & = \frac{ \min(n_t,p_k) }{p_k}, \label{eqn:coeff_rtk}
    \end{align}
    \begin{subnumcases}{\gamma_{t,k} \! = \!\!\label{eqn:coeff_gammatk}}
        \frac{\min(n_t,p_k)}{\max(n_t,p_k) \! - \! \min(n_t,p_k)\! - \!1}, \!\! &  $p_k \! \notin [n_t\pm 1]\! $ \\
        +\infty, \!\! & otherwise.
    \end{subnumcases}
where the notation $[n_t\pm 1]$ represents the range $[n_t \!-\! 1,\, n_t \!+\! 1]$.
Note that the infinity, i.e.,
$\infty$, is a short-hand notation for indeterminate/infinite values in the expressions.

\subsection{Generalization Error}

We now present our main result: 

\begin{theorem}\label{thm:expected_gen_isoG}
    Let Assumption~\ref{assume:standard_Gaussian_regressors} hold,
    and the noise vectors be independently distributed as 
    $\zvec_j\sim\Nc(\bm 0, \sigma_j^2\eye{n_j})$,
    and independent of the regressors. Let Assumption~\ref{assume:broad_and_large_Tc}
    or \ref{assume:T_c_is_one} hold.
    Then,
    over the distribution of $\what_t^{(T_c)}$
    as a function of the training data
    $\traindata{t} = \{(\yvec_j, \Amat_j)\}_{j=1}^t$,
    we have the expected generalization error in
    \eqref{eqn:expected_gen_err_def} as,    
    \begin{align}
    \begin{split}\label{eqn:gen_err_isoG}
        & G_T(\what_t^{(T_c)})
        = 
        \frac{1}{T}\sum_{i=1}^T
        \left\|\wvec_i^*\right\|^2_{\Hmatt{1}{t}} 
        + \phi\left(\what_t^{(T_c)}, \wvec_i^*\right)
        + \sigma_i^2
    \end{split}
    \end{align}
    where $\wvec_i^*$ denote the model unknowns in \eqref{eqn:model_matrix},
    and where
    \begin{align}\label{eqn:phi:result}
    \begin{split}
        \!\!
        \phi\!\left(\what_t^{(T_c)}\!\!\!, \wvec_i^*\right) \!\!
        & = \!\!\!
        \sum_{\tau=1}^t \! \bigg( \!\!
            \left\|\wvec_\tau^* \! - \! \wvec_i^*\right\|^2_{\Rmatt{\tau}{t}} \!\!
            + \! \sigma_\tau^2\frac{
                \sum_{k=1}^K\!\! \gamma_{\tau,k} h_{\tauplusone,k}}{K^2} \\
        & \qquad  + 2 \sum_{j=0}^{\tau-1}\left\langle
                \wvec_\tau^* - \wvec_i^*,
                \wvec_j^* - \wvec_i^*
            \right\rangle_{\Qmatt{\tau}{j}{t}} \bigg),
    \end{split}
    \end{align}
    with $\gamma_{t,k}$ defined as in \eqref{eqn:coeff_gammatk},
    and 
    where $\wvec_0^*=\bm 0$,
    \begin{align}
        \Hmatt{1}{t}     & = \diag{h_{1,k}\eye{p_k}}{k=1}{K} \\
        \Rmatt{\tau}{t} \!
            & = \! \diag{
                \frac{
                    h_{\tauplusone,k} r_{\tau,k} \!
                    + \!
                    \sum_{\substack{i=1\\i\neq k}}^K 
                        h_{\tauplusone, i}\gamma_{\tau, i}
                }{K^2} 
                \eye{p_k}
            }{k=1}{K}, \\
        \Qmatt{\tau}{j}{t} 
            & = \diagt{\Bigg\{
                \frac{r_{j,k}}{K} \!\!
                \prod_{\ell=j+1}^{\tau-1} \!\!
                \left( 1 \! - \! \frac{r_{\ell,k}}{K} \right)\!
                \\
            &\quad 
                \times
                \frac{h_{\tauplusone,k} r_{\tau,k} (K \!\! - \! 1)
                \! - \!\! \sum_{\substack{i=1\\i\neq k}}^K\!
                    h_{\tauplusone,i} \gamma_{\tau,i}}{K^2}\!
                    \eye{p_k}\!\!\Bigg\}
            }{k=1}{K}\!\!,
    \end{align}
    for $\tau=1,\,\dots,\,t$,
    and where $r_{t,k}$ is defined as in \eqref{eqn:coeff_rtk},
    with $r_{0,k} = K$,
    and where
    \begin{align}\label{eqn:h_tau_k_main_thm}
        h_{\tau,k} \! = \! \frac{
            h_{\tauplusone,k}\left(K^2 \! + \! r_{\tau,k} (1 \! - \! 2K)\right) 
            + \sum_{\substack{i=1\\i\neq k}}^K h_{\tauplusone,i} \gamma_{\tau,i} 
        }{K^2},
    \end{align}
    for $\tau=1,\,\dots,\,t$,
    and where $h_{t+1,k} = 1$, $\krange$.
    
\end{theorem}
\noindent Proof: See Appendix~\ref{proof:thm:expected_gen_isoG}.

    Theorem~\ref{thm:expected_gen_isoG} provides closed form expressions 
    for the generalization error for a stream of incoming tasks as a function of
    the noise levels $\sigma_i^2$,
    the number of samples per task $n_i$,
    the number of nodes in the network, $K$,
    and the number of unknowns governed by each node, $p_k$,
    as well as similarities between tasks through $\|\wvec_\tau^* - \wvec_i^*\|^2$ and   $\left\langle
                \wvec_\tau^* - \wvec_i^*,
                \wvec_j^* - \wvec_i^*
            \right\rangle$. 

\begin{remark}\label{remark:generalization:unseenTasks}
    For generality,
    we have used $\what_t^{(T_c)}$ where the estimate is trained on the first $t$ tasks,
    rather than $\what_T^{(T_c)}$ which is trained on all the $T$ tasks.
    As a result, the cross-terms 
    $\|\wvec_\tau^* - \wvec_i^*\|^2$ and
    $\left\langle
                \wvec_\tau^* - \wvec_i^*,
                \wvec_j^* - \wvec_i^*
    \right\rangle$, 
    i.e., the direct effect of task similarity, are only  present for $\tau \leq t$
    and $j<\tau$.
\end{remark}

\begin{remark}\label{remark:localstability}
    If $p_k \! \in [n_t\pm 1] $, 
    then the local sub-problems in the nodes become ill-conditioned with high-probability and the generalization error of \cocoa{} diverges.
    See \cite{hellkvist_ozcelikkale_ahlen_linear_2021} for detailed discussions. 
\end{remark}

    Theorem~\ref{thm:expected_gen_isoG} shows that   
    the network structure can have a large effect on the generalization error.  This effect goes beyond what has been described in Remark~\ref{remark:localstability}. 
    For instance, whether the error increases with the increasing number of tasks depends on the number of nodes in the network.
    The task similarity also heavily influences whether the error increases or decreases as the number of tasks increases.
    We further discuss these effects in Section~\ref{sec:num_similarity}.

\kern-0.5em
 \subsection{Special Case of Equal Dimensions}\label{sec:equalDimensions}

We now consider the special case where the problem dimensions are the same for all tasks and at all nodes:

\kern-0.1em
\begin{corollary}\label{res:gen:equaldimensions}
    Consider the setting of Theorem~\ref{thm:expected_gen_isoG}
    in the special case that
    all tasks have the same number of samples,
    i.e., $n_t = n$, $\forall t$,
    and all nodes have the same number of unknowns,
    and $p_k = \frac{p}{K}$,
    $\forall k$.
    With $r = r_{t,k}$ and $\gamma = \gamma_{t,k}$ as in 
    \eqref{eqn:coeff_rtk} -- \eqref{eqn:coeff_gammatk},
    we then have
    \begin{align}\label{eqn:gen:equaldimensions}
    \begin{split}
        & G_T\left(\what_t^{(T_c)}\right)
        = 
        \frac{1}{T}\sum_{i=1}^T \Bigg[
        \|\wvec_i^*\|^2 h^t
        + \sigma_i^2 \\
        & + \! \sum_{\tau=1}^t\! \bigg( \!
            \left\|\wvec_\tau^* - \wvec_i^*\right\|^2 
            \frac{
                r \! + \! (K \! - \! 1) \gamma
            }{K^2}h^{t-\tau} \!\!
            + \! \sigma_\tau^2\frac{
                \gamma}{K} h^{t-\tau} \\
        &   + 2 h^{t-\tau} 
            (r-\gamma)
            \frac{K-1}{K^2}
            \frac{r}{K}
            \sum_{j=1}^{\tau-1}
            \left(1 - \frac{r}{K}\right)^{\tau-j-1} \\
        & \qquad\qquad\qquad\qquad\qquad\qquad
            \times 
            \left\langle
                \wvec_\tau^* - \wvec_i^*,
                \wvec_j^* - \wvec_i^*
            \right\rangle \\
        & - 2\frac{K \! - \! 1}{K^2} \! 
            \left( \! 1  -  \frac{r}{K}\right)^{ \! \tau-1} 
            \!\! 
            (r-\gamma)
            h^{t-\tau}\!
            \left\langle
                \wvec_\tau^* - \wvec_i^*,
                \wvec_i^*
            \right\rangle\!
        \bigg)
        \Bigg],
    \end{split}
    \end{align}
    with
    \begin{equation}\label{eqn:h:equaldimensions}
        h = \frac{K^2 + (1-2K) r + (K-1)\gamma}{K^2}
    \end{equation}
    
\end{corollary}


We use Corollary~\ref{res:gen:equaldimensions} to provide comparisons with the centralized case  (Section~\ref{sec:comparison:centralized}) 
and to discuss the effect of task similarity (Section~\ref{sec:task-similarity} \galert{and Section~\ref{sec:ErrorDecomposition}}).

\subsection{Comparisons with the centralized continual learning}\label{sec:comparison:centralized}
Here we compare the results in  Corollary~\ref{res:gen:equaldimensions} with the expressions for the centralized continual learning setting of \cite{lin_2023_theory} in the overparametrized case, i.e. $p > n+1$.  
With only one node,
i.e., $K=1$ and $p_k=p$, the same noise level over all tasks, 
i.e., $\sigma_t^2 = \sigma^2$, focusing on the solution obtained after seeing the last task,
i.e., $t=T$,
and $T_c=1$,
then our scenario reduces to the setting of \cite{lin_2023_theory}.  
In
this setting,
\cocoa{} finds the convergence point of the stochastic gradient descent (SGD), i.e., the smallest-norm of the change of parameters  as in \cite[Eqn.4]{lin_2023_theory}. 
\balert{In particular, for this setting, \cocoa{} finds the solution of the  optimization problem
$ \min_{\what_t} \left\|\what_t - \what_{t-1} \right\|^2$  s.t.
$ \Amat_t \what_t = \yvec_t$, 
where $\what_{t-1}$  is the solution found for task $t-1$. Note that due to overparameterization, there are multiple solutions that satisfy $ \Amat_t \what_t = \yvec_t$. Hence,  \cocoa{} finds the solution that creates the minimum change in the parameter estimate while satisfying $ \Amat_t \what_t = \yvec_t$ for the new task $t$.}
As expected, the generalization error  in  Corollary~\ref{res:gen:equaldimensions} for $K=1$  matches that of \cite[Theorem~4.1]{lin_2023_theory},
i.e.,
\begin{align}
     \frac{(1-r)^T}{T}\sum_{i=1}^T \|\wvec_i^*\|^2 
     &+ \frac{1}{T}\sum_{\tau=1}^T r (1-r)^{T-\tau}
    \sum_{i=1}^T\left\|
        \wvec_\tau^* - \wvec_i^* 
    \right\|^2 \nonumber \\
    & + \frac{p \sigma^2}{p - n -1}(1-(1-r)^T),
\end{align}
where we have presented $G_T(\what_T^{(T_c)}) - \sigma^2$ to match the definition of the generalization error in \cite{lin_2023_theory}.
(Note that $r$ in \cite{lin_2023_theory} corresponds to $1-r$ according to our notation.)

\subsection{Task Similarity -- Single Model Parameter}\label{sec:task-similarity}
The generalization performance depends heavily on the similarity between tasks. 
We now investigate the generalization error for the limiting case of task similarity \galert{where there is one single model parameter for all tasks},  i.e. $\wvec_t^* = \wvec^*$  $\forall t$. 

\begin{theorem}\label{thm:similarity_generr}
    Consider the setting of Theorem~\ref{thm:expected_gen_isoG}.
    If $\wvec_t^* = \wvec^*$ and $\sigma_t^2 = \sigma^2$ $\forall t$,
    then 
    \begin{align}\begin{split}\label{eqn:gen_err_similar_tasks}
        \hspace{-0.75em}
        G_T\!\left(\!\what_T^{(T_c)}\!\right)\!
         & \! = \!
         \|\wvec^*\|^2_{\!\Hmatt{1}{T}} \!\!
            + \! \sigma^2 
            \frac{K^2 \!\! + \!\! \sum_{\tau=1}^T \!
            \sum_{k=1}^K \!\! \gamma_{\tau,k} h_{\tauplusone,k}}{K^2}.
    \end{split}
    \end{align}
    Furthermore, 
    if $\|\wvec^*\|^2 < \infty$ and for $t=1,\,\dots,\,T$, 
    \begin{align}\label{eqn:condition_n_smaller_than_pk}
    \begin{split}
       n_t
        < \pmin - \tfrac{K-1}{2K-1}\pmax - 1,
    \end{split}
    \end{align}
    or 
    \begin{align}\label{eqn:condition_n_larger_than_pk}
    \begin{split}
        n_t
        >  \pmax
        + \tfrac{K-1}{2K-1}\pmin
        + 1,
    \end{split}
    \end{align}
    where $\pmin = \min_{k=1,\dots,K} p_k$,
    $\pmax = \max_{k=1,\dots,K} p_k$,
    then    
    \begin{equation}\label{eqn:limit_H1_term}
        \lim_{T\to\infty} \|\wvec^*\|^2_{\Hmatt{1}{T}}
        = 0.
    \end{equation}
    
\end{theorem}
\noindent Proof: See Appendix~\ref{proof:thm:similarity_generr}.

\begin{corollary}\label{res:thm:similarity_generr:zeronoise}
Consider the setting of Theorem~\ref{thm:similarity_generr}
     with $\sigma^2 = 0$. Then, if \eqref{eqn:condition_n_smaller_than_pk} or \eqref{eqn:condition_n_larger_than_pk} is satisfied, the generalization error is zero in the limit of infinitely many tasks.
     
\end{corollary}
\noindent Proof:  The result follows from Theorem~\ref{thm:similarity_generr} with $\sigma^2=0$. 

\begin{remark}\label{remark:gen0conditions}
    Let $\sigma^2=0$.
    If $n_t \geq 1$ and $T\to\infty$,
    then the
    algorithm effectively processes an infinite number of samples,
    i.e., $n_t \times T \to\infty$. 
    Since the tasks have the same unknown parameter vector $\wvec^*\inrbb{p\times 1}$
    with $p < \infty$, one may expect that the generalization error should be zero unless the condition
    $p_k \! \notin [n_t \!-\! 1,\, n_t \!+\! 1]\! $ is violated, see Remark~\ref{remark:localstability}.
    Nevertheless, 
    Eqn.~\eqref{eqn:condition_n_smaller_than_pk}-\eqref{eqn:condition_n_larger_than_pk}  suggest stricter conditions on the problem dimensions may be needed. 
    Example~\ref{ex:h-diverges} illustrates this.
    
\end{remark}

\begin{example}\label{ex:h-diverges}
    We now illustrate Remark~\ref{remark:gen0conditions}. 
    In particular, we illustrate that Eqn.~\eqref{eqn:condition_n_smaller_than_pk}-\eqref{eqn:condition_n_larger_than_pk} provides non-trivial conditions for zero generalization error.
    Consider the equal dimensions scenario in Corollary~\ref{res:gen:equaldimensions} with $\sigma_t^2=0$, and $\wvec_t^* = \wvec^*$, $\forall t$. Let $n_t = 15$, $p = 40$ and $K=2$. Then $\gamma=15/4=3.75$, hence $p_k \! \notin [n_t \!-\! 1,\, n_t \!+\! 1]\! $ is satisfied but not \eqref{eqn:condition_n_smaller_than_pk} or \eqref{eqn:condition_n_larger_than_pk}. 
    Since $h$ of \eqref{eqn:h:equaldimensions} is $h =1.375$, evaluating \eqref{eqn:gen:equaldimensions} for $t=T$ shows that
    $h^T$  and hence the generalization error  $G_T(\what_T^{(T_c)})$ diverges  for $T \to \infty$.  
    
\end{example}

The next example illustrates that by only changing the network structure, one may have zero error instead of $G_T \to \infty$:

\begin{example}\label{ex:h-converges}
    Consider 
    Example~\ref{ex:h-diverges} 
    with
    $K=10$. 
    Then 
    $h = 0.846$ in \eqref{eqn:h:equaldimensions}, 
    hence $h^T\to 0$  and $G_T(\what_T^{(T_c)})\to 0$ as $T\to \infty$.
    
\end{example}

We also consider the limiting case of the training error:

\begin{lemma}
\label{col:forgetting_similarity}
    Consider the setting of Theorem~\ref{thm:similarity_generr}
    and $\sigma^2 = 0$.
    If \eqref{eqn:condition_n_smaller_than_pk} or 
    \eqref{eqn:condition_n_larger_than_pk}
    holds $\forall t$,
    and with the independence of $\Amat_t$ with the error $\what_T^{(T_c)}-\wvec^*$ 
    in the steady-state \cite[Ch. 16]{b_HaykinAdaptive},
    then
    \begin{equation}
        \lim_{T\to \infty}
        F_T(\what_T^{(T_c)}) 
        \approx 0.
    \end{equation}
    
\end{lemma}
\noindent Proof: See Appendix~\ref{proof:col:forgetting_similarity}.

To summarize, Corollary~\ref{res:thm:similarity_generr:zeronoise} and Lemma~\ref{col:forgetting_similarity} together show that 
if the local systems at the nodes are sufficiently away from a square system (with conditions stricter than the ones in Remark~\ref{remark:localstability}),
and the tasks are similar,
then 
both the generalization error and the training error can be made zero over all tasks.

\balert{\subsection{Task Similarity -- Decomposition of Error}\label{sec:ErrorDecomposition}}
\balert{%
We now investigate task similarity further by taking a closer look at different terms in the error expression. 
We start with the following intermediate result: 

\begin{corollary}\label{col:decomposition}
    Consider the setting of Corollary~\ref{res:gen:equaldimensions}.  Assume that 
    \begin{align}
        \sigma_i^2 & = \sigma^2, \\ 
        \left\|\wvec_i^*\right\|^2 & = E_1 ,\label{assumption:1_constant}\\
        \left\|\wvec_\tau^* - \wvec_i^*\right\|^2 & = E_2 \onebb_{i\neq \tau} ,\\
        \left\langle
            \wvec_\tau^* - \wvec_i^*,
            - \wvec_i^*
        \right\rangle
            & = E_3 \onebb_{i\neq \tau}, \\
        \left\langle
            \wvec_\tau^* - \wvec_i^*,
            \wvec_j^* - \wvec_i^*
        \right\rangle
            & = E_4 \onebb_{j\neq i\neq \tau \neq j}, 
            \label{assumption:4_constant}
    \end{align}
    where $i,\,\tau,\,j=1,\,\dots,\,T$,
    and $\onebb_{c(\cdot)} = 1$ if $c(\cdot)$ is true,
    and $0$ otherwise.
    %
    Then,
    the expected generalization error for $t=T$
    in \eqref{eqn:gen:equaldimensions}
    can be rewritten as
    \begin{align} \label{eqn:GT_psi_i_U_i}
        G_T\!\left(\what_T^{(T_c)}\right)
        \!= \psi_0 \sigma^2
        + \psi_1 E_1
        + \psi_2 E_2
        + \psi_3 E_3
        + \psi_4 E_4,
    \end{align}
    where
    \begin{align}
        \psi_0 & = 1 + \frac{\gamma}{K}\frac{1 - h^T}{1 - h}, \label{coeff:noise} \\
        \psi_1 & = h^T, \\
        \psi_2 & = \frac{r+(K-1)\gamma}{K^2}\frac{T-1}{T}\frac{1-h^T}{1-h}, \label{coeff:psi2}\\
        \psi_3 & = 2 \frac{(K-1)(r-\gamma)}{K^2}\frac{T-1}{T}\frac{h^T-b^T}{h-b},\label{coeff:psi3} \\
        \psi_4 & = 2 \frac{(K-1)(r-\gamma)}{K^2} \frac{T-2}{T} \left( \frac{1-h^T}{1-h} - \frac{b^T-h^T}{b-h} \right),\label{coeff:psi4} 
    \end{align}
    for $h \neq 1$
    and $b\neq h$,
    where $h$ is defined in \eqref{eqn:h:equaldimensions}
    and $b = 1 - \frac{r}{K}$.   
    %
    If $K=1$ and $r=1 $,
    then $h=b=0$,  and $\psi_0$-$\psi_2$ are given by \eqref{coeff:noise}-\eqref{coeff:psi2}, and  $\psi_3 = \psi_4 = 0$.
\end{corollary}

\noindent Proof: The result follows from Corollary~\ref{res:gen:equaldimensions} with algebraic manipulations.

In order to systematically study the task similarity, we introduce the following  model for the  task parameters: 

\begin{model}\label{model:sharedps}
    Let $0 \leq p_S \leq p$. The unknown vectors $\wvec_t^*$ share the first $p_S$ entries, 
    \begin{equation}\label{eqn:model:sharedps}
        \wvec_t^* = 
        \begin{bmatrix}
        \Bar{\wvec}^* \\ \Bar{\wvec}_t^*
        \end{bmatrix},
    \end{equation}
    where
    $\Bar{\wvec}^*\inrbb{p_S\times 1}$ is the same for all tasks,
    and $\Bar{\wvec}_t^*\inrbb{(p-p_S)\times 1}$ is specific to each task $t$. 
    Here, $\Bar{\wvec}^*$,  and
    $\Bar{\wvec}_1^*,\, \ldots,\, \Bar{\wvec}_T^*$,
    are zero-mean and statistically independent.   
    Hence,
    $
       \eunder{\Wc}\left[ 
            (\wbar_i^*)\T \wbar_\tau^*
        \right] = 0, \, i\neq \tau,
    $
    where $\Wc$ denotes the joint probability distribution function of the parameters,
    i.e., $(\wbar^*,\wbar_1^*,\,\dots,\,\wbar_T^*)\sim \Wc$.
    The covariance matrices of $\Bar{\wvec}^*$ and $\Bar{\wvec}_t^*$ are given by $ \sigma_{w}^2\, \eye{p_S}$ and $\sigma_{w}^2 \eye{p-p_S}$, respectively.
    Hence,  the variable $p_S$,
    which specifies the number of shared unknowns between the tasks and
    controls the relative power of the shared component $\Bar{\wvec}^* $, is
    providing a measure of the task similarity for a fixed $p$.
    We have 
    \begin{align}
        \label{assumption:1_constant:modelps}
          \eunder{\Wc} \left[   \left\|\wvec_i^*\right\|^2 \right]& = E_w\\
           \eunder{\Wc} \left[    \left\|\wvec_\tau^* - \wvec_i^*\right\|^2  \right] & = 2 \frac{p \! - \! p_S}{p} E_w, \, {i\neq \tau} ,\\
         \eunder{\Wc} \left[    \left\langle
                \wvec_\tau^* \!- \!\wvec_i^*,
                - \wvec_i^*
            \right\rangle \right]
                & =   \frac{p \! - \! p_S}{p} E_w ,\, {i\neq \tau}, \\
         \eunder{\Wc} \left[
            \left\langle
                \wvec_\tau^* \! - \! \wvec_i^*,
                \wvec_j^* \! - \! \wvec_i^*
            \right\rangle \right]
                & = \frac{p \! - \! p_S}{p} E_w ,\, {j\neq i\neq \tau \neq  j},
                \label{assumption:4_constant:modelps}
        \end{align}
    where $E_w = p  \sigma_{w}^2$.
\end{model}

\begin{corollary}\label{col:decomposition:modelps}
    Under the setting of Corollary~\ref{res:gen:equaldimensions}, 
    the expected value of the generalization error under Task Model~\ref{model:sharedps},
    i.e.,
    $\eunder{\Wc}[
        G_T\!\left(\what_T^{(T_c)}\right)
    ]$,  
    is given by  
    \begin{align}
        \psi_0\sigma^2 + \left(\psi_1 + \frac{p-p_S}{p}\left(2\psi_2 + \psi_3 + \psi_4 \right)\right) E_w.
    \end{align}
  
\end{corollary}

\noindent Proof: The result follows by taking the expectation of both sides of \eqref{eqn:gen:equaldimensions} with respect to $\Wc$, which results in an expression in the form of \eqref{eqn:GT_psi_i_U_i} where  
$E_1$ -- $E_4$ in
\eqref{assumption:1_constant} -- \eqref{assumption:4_constant} are substituted with the values in  
\eqref{assumption:1_constant:modelps} -- \eqref{assumption:4_constant:modelps}.

We now illustrate the  performance  under Task Model~\ref{model:sharedps} for different  $K$ and $T$ values. 
For a  \cocoa{} solution $\what$ obtained with  $K=\bar{K}$ nodes,  we denote the error as 
\begin{align}
    \bar{G}_{T}^{\bar{K}} =\eunder{\Wc}\left[ 
            G_T\!\left(\what_T^{(T_c)}\right)
        \right],
\end{align}
with the convention
$\bar{G}_{\infty}^{\bar{K}} \!=\!\lim_{T\to\infty} \eunder{\Wc}\left[
        G_T\!\left(\what_T^{(T_c)}\!\right)
    \right]$. 
We have 
\begin{align}\label{eqn:modelsharedps:largeT}
    \bar{G}_{\infty}^{\bar{K}} 
    = \left(1 + \frac{\gamma}{K}\frac{1}{1 - h}\right) \sigma^2
    + \frac{2 r}{K}\frac{1}{1 - h} \left( \frac{p-p_S}{p} E_w\!\right),
\end{align}
under $|h|< 1$.
The condition $|h|< 1$ is satisfied for a 
wide range of $p$, $n_t$, $K$ combinations,
e.g., under the conditions in \eqref{eqn:condition_n_smaller_than_pk}
and \eqref{eqn:condition_n_larger_than_pk}.
One such scenario is presented in Example~\ref{ex:G_for_n=2p_T=inf}.
Considering $\bar{G}_{\infty}^{\bar{K}} -\sigma^2$, we observe that 
both the noise level ($\sigma^2$) and the average power of the non-shared part of task unknown vectors 
($({(p-p_S)}/{p}) \times E_w$) is scaled with a coefficient in the form of
${\alpha }/{(K (1 - h))}$ where $\alpha =\gamma$  and $\alpha =2 r$ for the noise and signal  components, respectively. 

The following example illustrates that in the underparametrized case, \cocoa{} can provide lower error than the online centralized continual learning solution for large $T$.

\begin{example}\label{ex:G_for_n=2p_T=inf}
     Consider the setting of Corollary~\ref{col:decomposition:modelps} with
     $n_t = 2p$,
    $p\geq 2$,
    $T_c=1$.
    Under Task Model~\ref{model:sharedps},  we have  
    \begin{align}\label{eqn:G_for_n=2p_limit_K=1}
        \bar{G}_{\infty}^{1}
         =\left(1 \! + \! \frac{p}{p-1}\right) \sigma^2
        + \!\left(2 \frac{p-p_S}{p} E_w\!\right),
    \end{align}
    \begin{align}\label{eqn:G_for_n=2p_limit_K=p}
      \bar{G}_{\infty}^p 
        \!= \!\left( 1 \! + \! \frac{p}{4p-3}\frac{1}{p-1} \right) \sigma^2
        \!+\! \frac{2p}{4p-3} 
       \!\left(2 \frac{p-p_S}{p} E_w\!\right),
    \end{align}
    which have been obtained by inserting the values of $r$, $\gamma$ and $h$
    determined by $n_t$, $p$ and $K$ used here.
    The expressions $\bar{G}_{\infty}^{1}$ in \eqref{eqn:G_for_n=2p_limit_K=1}
    and $\bar{G}_{\infty}^p$ in \eqref{eqn:G_for_n=2p_limit_K=p}
    give the expected generalization error under Task Model~\ref{model:sharedps}
    for the centralized continual learning setting $K=1$
    and the distributed setting with $K=p$,
    respectively.
    Comparing the error in
    \eqref{eqn:G_for_n=2p_limit_K=p},
    i.e., for $K=p$, 
    and the error in \eqref{eqn:G_for_n=2p_limit_K=1},
    i.e., for $K=1$,
    we observe that the error is lower
    in the distributed setting of $K=p$
    than in the centralized setting of $K=1$,
    regardless of the value of $p_S$. 
    We also note that when there is no noise, 
    i.e., $\sigma^2=0$,
    the error under $K=p$ converges to $\approx \frac{p-p_S}{p} E_w$ when $T\to\infty$, 
    which is the average power of the non-shared part of the task parameters.
    Hence, \cocoa{} finds the shared part of the task unknowns despite its iterative and distributed nature.

\end{example}

\begin{figure}
     \begin{subfigure}[b]{0.49\linewidth}
         \centering
         \includegraphics[width=0.89\textwidth]{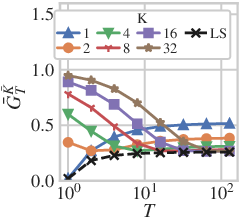}
         \caption{ \balert{$\sigma^2 = 0.01$ (high SNR)}}
         \label{fig:ex3:p32:n64:highSNR}
     \end{subfigure}
     \begin{subfigure}[b]{0.49\linewidth}
         \centering
         \includegraphics[width=0.89\textwidth]{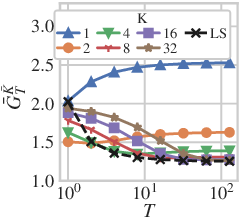}
         \caption{\balert{$\sigma^2 = 1$ (low SNR)}}
         \label{fig:ex3:p32:n64:lowSNR}
     \end{subfigure}
        \caption{\balert{The expected error $\bar{G}_{T}^{\bar{K}}$ versus $T$ under Task Model~\ref{model:sharedps},
        with $p=32$, $p_S = 24$, $n_t=64$, $T_c=1$, and $E_w=1$.}}
        \label{fig:ex3:p32:n64}
\end{figure}
\begin{figure}
     \begin{subfigure}[b]{0.49\linewidth}
         \centering
         \includegraphics[width=0.89\textwidth]{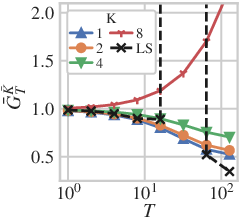}
         \caption{\balert{$\sigma^2 = 0.01$ (high SNR)}}
         \label{fig:ex3:p32:n1:highSNR}
     \end{subfigure}
     \begin{subfigure}[b]{0.49\linewidth}
         \centering
         \includegraphics[width=0.89\textwidth]{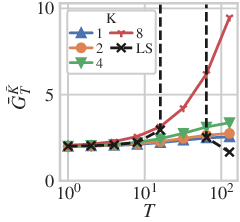}
         \caption{\balert{$\sigma^2 = 1$ (low SNR)}}
         \label{fig:ex3:p32:n1:lowSNR}
     \end{subfigure}
        \caption{\balert{The expected error $\bar{G}_{T}^{\bar{K}}$ versus $T$ under Task Model~\ref{model:sharedps},
        with $p=32$, $p_S = 24$, $n_t=1$, and $E_w=1$.}}
        \label{fig:ex3:p32:n1}
\end{figure}

In Fig.~\ref{fig:ex3:p32:n64} and Fig.~\ref{fig:ex3:p32:n1}, 
we present $\bar{G}_{T}^{\bar{K}}$ versus $T$ under Task Model~\ref{model:sharedps} for $p=32$, $p_S = 24$,
for the underparametrized and overparametrized scenarios, respectively.
The curve with the legend ``LS'' 
plots the error of the least-squares (LS) benchmark presented in Section~\ref{sec:offlineCentralized} 
and is obtained numerically,
whereas the other curves are plotted using Corollary~\ref{col:decomposition:modelps}.
We note that the high error values for the LS solution around $T=32$ in Fig.~\ref{fig:ex3:p32:n1}
is due to the ill-conditioning of the system for $T n_t \approx p$, similar to the scenarios in Remark~\ref{remark:localstability}.
Fig.~\ref{fig:ex3:p32:n64} illustrates that the \cocoa{} solutions can achieve error values that are close to that of the LS solution for large $T$ and large $K$; and can even obtain error values lower than the LS solution for small $T$ under low SNR (Fig.~\ref{fig:ex3:p32:n64:lowSNR}).    
In Fig.~\ref{fig:ex3:p32:n1}, 
we observe that the error values of \cocoa{} are  close to those of the LS solution for small and moderate $T$ ($1 \lessapprox T \lessapprox 10$)  for most $K$.
Fig.~\ref{fig:ex3:p32:n1} illustrates that the error of \cocoa{}
for large $T$ ($T \gtrapprox 16$) can decrease or increase with increasing $T$ depending on the noise level and $K$,
although the lowest error values for the LS solution across different $T$  are obtained with the large value of $T \approx 100$. 

}

%% file: numerical_results.tex
\section{Numerical Results}\label{sec:numericalResults}
We now present numerical results in order to to illustrate the continual learning performance of \cocoa{} and the analytical results of  Section~\ref{sec:gen_err_analytical}.

We include an experiment with real-world data in Section~\ref{sec:num:MNIST} whose details are explained therein. In the rest of the experiments, we use the following setting:  
For each experiment,
we generate a set of task unknowns
$\wvec_t^*$, $\forall t$.
In order to control the task similarity,  \galert{we use the variable $p_S$, as in \eqref{eqn:model:sharedps},  where}
 the unknown vectors $\wvec_t^*$ share the first $p_S$ entries,
where
$\Bar{\wvec}^*\inrbb{p_S\times 1}$ is the same for all $\wvec_t^*$,
and $\Bar{\wvec}_t^*\inrbb{(p-p_S)\times 1}$ is independently generated for each $t$.
We draw $\Bar{\wvec}^*$ and $\Bar{\wvec}_t^*$
from the standard Gaussian distribution,
and normalize the drawn vectors such that $\|\Bar{\wvec}^*\|^2 = \frac{p_S}{p}$
and $\|\Bar{\wvec}_t^*\|^2 = \frac{p-p_S}{p}$.
Hence, $\|\wvec_t^*\|^2 = 1$.

The entries of the regressors matrices $\Amat_t$ are generated  i.i.d. from $\Nc(0, 1)$,
hence Assumption~\ref{assume:standard_Gaussian_regressors} is fulfilled.
The noise vectors $\zvec_t$ are independently drawn from $\Nc(0, \sigma_t^2\eye{n_t})$,
where $\sigma_t^2$ may vary over different experiments.
The observations $\yvec_t$
are generated as in \eqref{eqn:model_matrix}, i.e., $\yvec_t = \Amat_t\wvec_t^*+\zvec_t$.
We have $p_k = \frac{p}{K}$, $\forall k$. %
With $\what_0^{(0)}=\bm 0$,
we run Algorithm~\ref{alg:continual_cocoa} for $\trange$ to obtain the estimate $\what_T^{(T_c)}$, 
trained over all tasks.
We then compute the generalization error
as $g_T(\what_T^{(T_c)})$ in \eqref{eqn:gen_err_def},
and create an average over $100$ i.i.d. sets of training data to report the empirical value of $G_T(\what_t^{(T_c)})$.

\subsection{Verification of Theorem 1 and Effect of Network Size}
\kern-0.25em

    \label{sec:num_verification_network_size}
    \begin{figure}
    \begin{subfigure}{0.49\linewidth}
        \includegraphics[width=1\textwidth]{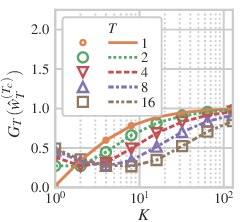}
        \caption{
        $n_t\!=\!2048$,
        $\sigma_t^2\!=\!0.01$,
        $T_c \!=\! 1$.}
        \label{fig:G_vs_K}
    \end{subfigure}
    \begin{subfigure}{0.49\linewidth}
        \includegraphics[width=1\textwidth]{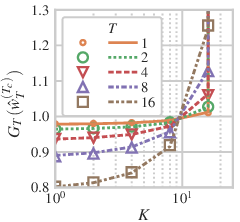}
        \caption{
        $n_t\!=\!32$,
        $\sigma_t^2\!=\!0.01$,
        $T_c \!=\! 100$.}
        \label{fig:G_vs_K_overp}

    \end{subfigure}
    \caption{The generalization error versus the number of nodes, 
        for different number of tasks (marker: analytical, line: simulations).
        Here, $p=1024$,
        $p_S = 768$.}
    \label{fig:G_vs_K_both}
    \end{figure}
    
    In Fig.~\ref{fig:G_vs_K_both}
    we plot the generalization error $G_T(\what_T^{(T_c)})$ 
    versus the number of nodes in the network $K$,
    for different numbers of tasks $T$.
    The x-axis shows the number of nodes $K$ and
    is sampled at $\{2^0,~2^1,~2^2,\, \dots\}$.
    For both figures, $p=1024$, 
    $p_S=768$ and 
    $\sigma_t^2=0.01$.
    In Fig.~\ref{fig:G_vs_K},
        $n_t=2048$ and $T_c = 1$,
    thus Assumption~\ref{assume:T_c_is_one} is fulfilled,
    i.e., $T_c = 1$. 
    In Fig.~\ref{fig:G_vs_K_overp},
        $n_t=32$ and       
        $T_c = 100$,
    thus Assumption~\ref{assume:broad_and_large_Tc} is fulfilled for $K < 32$,
    i.e., $p_k > n_t+1,~\forall t, k $.

    We plot the average generalization error 
    obtained via simulations (lines),
    together with the analytically evaluated expected generalization error by Theorem~\ref{thm:expected_gen_isoG}
    (markers).
    We observe that analytical and empirical results match both in Fig.~\ref{fig:G_vs_K} and \ref{fig:G_vs_K_overp}, 
    i.e., either under Assumption~\ref{assume:broad_and_large_Tc} or Assumption~\ref{assume:T_c_is_one}. 

    We observe that the number of nodes $K$
    has a non-trivial effect on the generalization error.
    For instance, in Fig.~\ref{fig:G_vs_K}, 
    the error first decreases (for $T\geq2$)
    as the number of nodes $K$ increases,
    and then increases to approach 
    $\frac{1}{T}\sum_{t=1}^T\|\wvec_t\|^2 + \sigma_t^2 = 2$, which is the generalization error of the zero-estimator $\what=\bm 0$.
    The behaviour with $K$
    is closely related to the task similarity,
    which is discussed in 
    Section~\ref{sec:num:taskSimilarity}.
    The generalization error in Fig.~\ref{fig:G_vs_K_overp} increases gradually 
    with $K$ up to $K=16$,
    where the error starts to increase rapidly due 
    to local ill-conditioning
    for $p_k \approx n_t$,
    see Remark~\ref{remark:localstability}.

    We now compare generalization performance  of \cocoa{} with the  performance of
    the offline and centralized solution 
    $\offlinesoln$ in \eqref{eqn:offline_centralized_solution}.
    The error associated with
    $\offlinesoln$ in \eqref{eqn:offline_centralized_solution}
    for $T = 1, 2, 4, 8, 16$,
    is given by $\approx \{0.02, 0.18, 0.22, 0.25, 0.25\}$  for  
    Fig.~\ref{fig:G_vs_K} and $\approx \{0.98, 0.97, 0.94, 0.90, 0.87\}$  for Fig.~\ref{fig:G_vs_K_overp}.  
    We observe that in some but not all scenarios, by tuning the number of nodes 
    $K$ for a given number of tasks $T$, one may obtain close or even lower values of expected generalization error with \cocoa{}
    in comparison to the offline and centralized solution $\offlinesoln$.
    For instance,
    in Fig.~\ref{fig:G_vs_K_overp} for $T=16$ and $K = 2$
    the \cocoa{} error is $0.81$, which is lower than 
    $0.87$ for $\offlinesoln$.

    We observe that the error behaviour with 
    the number of tasks 
    $T$ depends on the number of nodes $K$,
    and is affected by the tasks' similarities,
    which is investigated in Section~\ref{sec:num:taskSimilarity}.

\subsection{Generalization Error and Task Similarity}\label{sec:num:taskSimilarity}
    
\label{sec:num_similarity}
    \begin{figure}
        \centering
        \includegraphics[width =0.85 \linewidth]{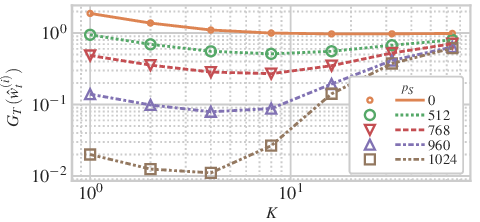}
        \caption{The generalization error versus the number of nodes, 
        for different number of shared task unknowns $p_S$,
        i.e., task similarity (marker: analytical, line: simulations).
        Here, $n_t=2048$,
        $p=1024$,
        $\sigma_t^2=0.01$,
        $T=16$,
        $T_c = 1$.}
        \label{fig:G_vs_K_similarity}

    \end{figure}
    
    \begin{figure}
        \begin{subfigure}{0.49\linewidth}
            \centering
            \includegraphics[width=1\textwidth]{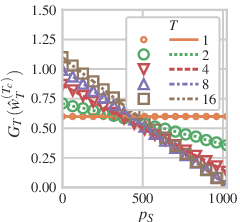}
            \caption{$K=4$}
            \label{fig:G_vs_pS_similarity:Klow}
        \end{subfigure}
        \begin{subfigure}{0.49\linewidth}
            \centering
            \includegraphics[width=1\textwidth]{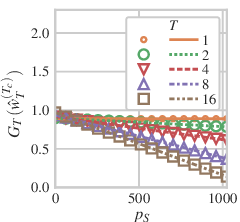}
            \caption{$K=16$}
            \label{fig:G_vs_pS_similarity:Khigh}
        \end{subfigure}
         \caption{The generalization error versus the number of shared task unknowns $p_S$,
         i.e., task similarity (marker: analytical, line: simulations).
         Here, $n_t=2048$,
         $p=1024$,
         $\sigma_t^2=0.01$,
        $T_c = 1$.}
        \label{fig:G_vs_pS_similarity}
    \end{figure}
    In Fig.~\ref{fig:G_vs_K_similarity},
    we plot the generalization error 
    versus the number of nodes $K$
    for different levels of similarity,
    i.e., number of shared parameters $p_S$.
      Here,
    $n_t = 2048$,
    $p=1024$,
     $\sigma_t^2 = 0.01$,
     $T=16$,
    and $T_c=1$,
    thus Assumption~\ref{assume:T_c_is_one} is fulfilled.
    We again observe a match between the analytical and simulated error curves as we vary $p_S$ and $K$.
    This figure quantifies the possible large impact of task similarity on the generalization performance; and the dependence of this impact on the network size $K$.
    Compared to having no shared parameters between the tasks ($p_S=0$),
    more similarity (larger $p_S$) decreases the generalization error.
    In particular,
    for $p_S = 1024$, i.e. the parameter vector is the same for all tasks, 
    the error for $K=4$ is close to the noise floor of $\sigma_t^2 = 0.01$.
    For large $K$, the increase in the error with increasing $K$ is consistent with the fact that with too many nodes, the number of possibly incompatible local estimates becomes too high, hence  more rounds of communications may be needed to reach compatibility between these large number of local estimates. 

    We plot the  generalization error versus the number of shared parameters $p_S$
    with varying numbers of tasks $T$ for $K=4$ and  $K=16$ in   Fig.~\ref{fig:G_vs_pS_similarity:Klow} and Fig.~\ref{fig:G_vs_pS_similarity:Khigh}, respectively. 
    These plots quantify how for dissimilar tasks, 
    i.e., small $p_S$,
     the error increases with the number of tasks $T$,
    and for similar tasks, i.e., large $p_S$,
    the error decreases as the number of tasks increases.
    This is consistent with the fact that it may be possible to obtain good estimates with a model with a single parameter vector if the tasks are sufficiently similar;
    and that with a larger $T$,
    the effective number of observations seen by $\cocoa{}$ increases,
    which then can be used to estimate this single parameter vector. 
    Comparing Fig.~\ref{fig:G_vs_pS_similarity:Klow} and Fig.~\ref{fig:G_vs_pS_similarity:Khigh}, 
    we observe that the network size $K$ affects the range of $p_S$ on which the curves cross,
    i.e.,
    for which $p_S$ a larger number of tasks $T$ is beneficial.
    For a large number of nodes $K$ (Fig.~\ref{fig:G_vs_pS_similarity:Khigh}),
    continual learning with $T>1$ with tasks that are even dissimilar,
    i.e. small $p_s$,
    can provide relatively good performance compared to the case with $T=1$.

\subsection{Learning Curves}
    \begin{figure}
     \begin{subfigure}[b]{0.47\linewidth}
         \centering
         \includegraphics[width=1\textwidth]{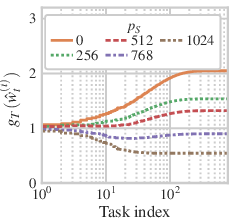}
         \caption{}
         \label{fig:G_and_F_vs_t_overp:generalization}
     \end{subfigure}
     \begin{subfigure}[b]{0.49\linewidth}
         \centering
         \includegraphics[width=1\textwidth]{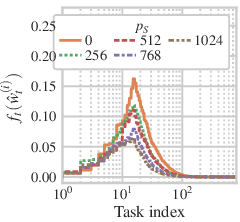}
         \caption{}
         \label{fig:G_and_F_vs_t_overp:training}
     \end{subfigure}
        \caption{The generalization error (a) and training error (b)
        for the estimate $\what_t^{(i)}$ versus the task index,
        i.e., seen tasks,
        for different numbers of shared unknowns $p_S$.
        Here, $n_t=32$,
        $p=1024$,
        $\sigma_t^2=0.01$,
        $K=2$,
        $T_c=100$,
        $T=16$.
        }
        \label{fig:G_and_F_vs_t_overp}
    \end{figure}
    
    \begin{figure}
     \begin{subfigure}[b]{0.49\linewidth}
         \centering
         \includegraphics[width=1\textwidth]{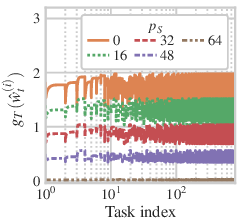}
         \caption{}
         \label{}
     \end{subfigure}
     \begin{subfigure}[b]{0.49\linewidth}
         \centering
         \includegraphics[width=1\textwidth]{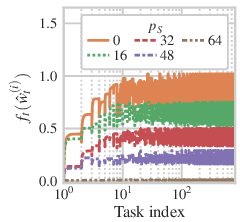}
         \caption{}
         \label{}
     \end{subfigure}
        \caption{The generalization error (a) and training error (b)
        for the estimate $\what_t^{(i)}$ versus the task index,
        i.e., seen tasks,
        for different numbers of shared unknowns $p_S$.
        Here, $n_t=128$,
        $p=64$,
        $\sigma_t^2=0.01$,
        $K=2$,
        $T_c=100$,
        $T=16$.
        }
        \label{fig:G_and_F_vs_t_underp}
    \end{figure}

    In Fig.~\ref{fig:G_and_F_vs_t_overp} and \ref{fig:G_and_F_vs_t_underp}, 
    we study the evolution of the generalization error and the training error
    as the the model is repeatedly trained on the tasks $\trange$. In particular, we plot 
    $g_T(\what_t^{(i)})$ and $f_t(\what_t^{(i)})$,
    see \eqref{eqn:gen_err_def} and \eqref{eqn:forgetting_def},
    respectively.
    The estimate $\what_T^{(T_c)}$ 
    does not necessarily converge after training on all tasks once,
    hence we repeat the tasks after all $T$ tasks have been trained upon.
    The resulting task sequence is
    $t'=1,\,\dots,\,T,\,1,\,\dots,\,T,\dots$,
    where $t'$ corresponds to the task index on the x-axes in
    Fig.~\ref{fig:G_and_F_vs_t_overp} and \ref{fig:G_and_F_vs_t_underp}.
    For each task index $t'$,
    we plot the error over the $T_c$ iterations of \cocoa{},
    hence there are $T_c$ points
    in between every integer task index.

    In Fig.~\ref{fig:G_and_F_vs_t_overp},
     $T_c=100$, $\sigma_t^2 = 0.01$, $T=16$,  $n_t = 32$, $p=1024$, $K=2$.
    Hence,
    $p_k = 512 > n_t = 32$,
    and Assumption~\ref{assume:broad_and_large_Tc} is fulfilled
    and Lemma~\ref{res:one-step-convergence} holds,
    i.e., \cocoa{} converges in the first iteration. 
    Accordingly, in Fig.~\ref{fig:G_and_F_vs_t_overp},
    the generalization and training error are
    constant over the \cocoa{} iterations between each integer task index.
    Furthermore, these plots provide insights into the interplay between the level of task similarity and the number of tasks.
    We observe that if the similarity is relatively low,
    e.g., for $p_S=0,256,\,512$,
    then the generalization error in Fig.~\ref{fig:G_and_F_vs_t_overp:generalization} increases for each task trained upon.
    On the other hand,
    the training error in Fig.~\ref{fig:G_and_F_vs_t_overp:training} decreases as the task index increases after $T=16$,
    i.e., after all tasks have been trained on once.
    This effect is consistent with overfitting to training data under the locally overparameterized setting of $p_k = 512 > n_t = 32$, i.e.,  
    the model becomes good at predicting the seen samples of data,
    while generalizing poorly.

    In Fig.~\ref{fig:G_and_F_vs_t_underp},
    we repeat the experiment of Fig.~\ref{fig:G_and_F_vs_t_overp},
    but  $n_t=128$ and $p=64$.
    As $K=2$,
    we have $p_k < 128$,
    hence Lemma~\ref{res:one-step-convergence} does not hold,
    and we observe a learning transient for each seen task,
    i.e., the error curves between each integer task index.
    After training on all $T=16$ tasks,
    both the generalization and training errors 
    fluctuate around fixed mean values,
    and the mean generalization error for a given $p_S$ improves as $p_S$ increases.

    \subsection{Effect of the Number of Iterations $T_c$}\label{sec:num:Tc}
\begin{figure}
        \centering
        \includegraphics[width=0.81\linewidth]{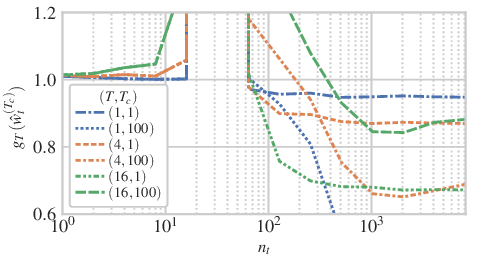}
        \caption{
        The generalization error versus number of samples per task $n_t$,
        for different number of tasks $T$ and number of iterations $T_c$ of \cocoa{}.
        Here,
        $K=32$,
        $p=1024$,
        $p_S = 512$,
        $\sigma_t^2=0.01$.}
        \label{fig:G_vs_nt}
    \end{figure}

    In Fig.~\ref{fig:G_vs_nt},
    we plot the generalization error
    versus the number of samples per task $n_t$
    for different number of tasks $T$,
    for both $T_c=1$ and $T_c = 100$ iterations per task.
    Here, $\sigma_t^2 = 0.01$, 
    $K=32$,
    $p=1024$, $p_S=512$. 

    Since $p_k = \frac{p}{K} = 32$,
    for $n_t < 32$ the curves for $T_c=1$ and $T_c=100$ are on top of each other due to the convergence of \cocoa{} in the first iteration $i=1$, see Remark~\ref{rem:iso-G-large_Tc}.
    As $n_t$ increases toward $n_t=32$,
    there is a large peak in the error for $p_k \approx n_t$,
    since the local problems are now at the interpolation threshold, 
    hence ill-conditioned, see Remark~\ref{remark:localstability}.
    
    For $n_t\geq 32$, 
    the error decreases again with increasing $n_t$,
    and the curves for $T_c=1$ and $T_c=100$ exhibit different behaviour.
    Here,
    the behaviour is different between the pairs of curves for different number of tasks $T$.
    In particular,
    if $T=1$,
    then the error is lower with high $T_c$,
    and if there are many tasks, 
    i.e., if $T=16$,
    the relation is flipped and the error is higher with higher $T_c$.
    Hence,
    if there are many tasks with not enough task similarity,
    it may be beneficial in terms of generalization error to run \cocoa{} for a smaller number of iterations, i.e., apply early stopping, 
    for each task.

    \subsection{Effect of Feature Correlation}\label{sec:num:correlation}
\begin{figure}
    \begin{subfigure}{0.49\linewidth}
        \includegraphics[width=1\textwidth]{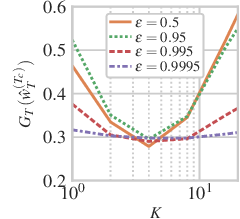}
        \caption{
        $n_t\!=\!2048$,
        $\sigma_t^2\!=\!0.01$,
        $T_c \!=\! 1$.}
        \label{fig:G_vs_K:corr}
    \end{subfigure}
    \begin{subfigure}{0.49\linewidth}
        \includegraphics[width=1\textwidth]{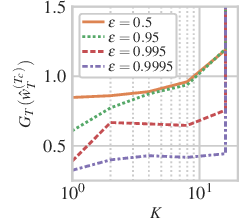}
        \caption{
        $n_t\!=\!32$,
        $\sigma_t^2\!=\!0.01$,
        $T_c \!=\! 100$.}
        \label{fig:G_vs_K_overp:corr}
    \end{subfigure}
    \caption{\galert{The generalization error versus the number of nodes $K$, 
        for different levels of correlation.
        Here, $p=1024$,
        $p_S = 768$,
        $T=8$.}}
    \label{fig:G_vs_K_both:corr}
    \end{figure}

\galert{%
   We now investigate the effect of feature correlation. The covariance matrices of the regressors are symmetric Toeplitz matrices where the first row is given by $[\varepsilon^0,\varepsilon^1, \ldots  \varepsilon^{(p-1)}]$ where $0 \leq \varepsilon \leq 1$.   Hence, increasing values of $\varepsilon$ results in higher values of correlation between features.  
   We consider the setting in Fig.~\ref{fig:G_vs_K_both} with $T=8$.  
   The resulting plots are provided in Fig.~\ref{fig:G_vs_K_both:corr}. 
   In general, lower error values are obtained for higher values of correlation although exceptions to this trend exist, e.g.,  Fig.~\ref{fig:G_vs_K:corr} for $\varepsilon=0.5, 0.95$ and small $K$. 
   Similar to Fig.~\ref{fig:G_vs_K_overp}, the error in Fig.~\ref{fig:G_vs_K_overp:corr} rapidly increases with $K=16$ where we get closer to the case of $p_k \approx n_t$; highlighting the importance of  Remark~\ref{remark:localstability} also under correlated regressors.  
}%
    
\subsection{Continual Learning on MNIST with \cocoa{}}\label{sec:num:MNIST}
    \begin{figure}
        \centering
        \includegraphics[width =0.78 \linewidth]{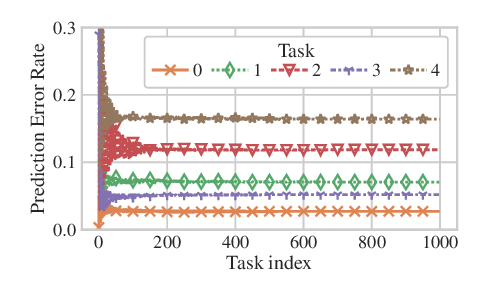}
        \caption{The test prediction error rate for the odd/even MNIST classification task when trained with  quadratic loss,
        versus the number of repetitions of Algorithm~\ref{alg:continual_cocoa}.
        Here, $K=2$,
        $p=3\cdot 10^3$,
        $n_t=100$,
        and $T_c = 1$.}
        \label{fig:mnist}
    \end{figure}
    \begin{figure}
        \centering
        \includegraphics[width =0.78 \linewidth]{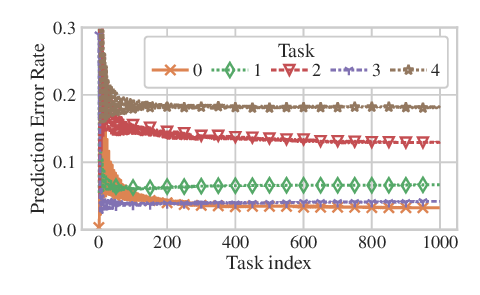}
        \caption{ \balert{The test prediction error rate for the odd/even MNIST classification task when trained with logistic loss,
        versus the number of repetitions of Algorithm~\ref{alg:continual_cocoa}.
        Here, $K=2$,
        $p=3\cdot 10^3$,
        $n_t=100$,
        and $T_c = 1$.}}
        \label{fig:mnist:logistic}
    \end{figure}

     We now study the continual learning performance of \cocoa{} using the MNIST dataset \cite{lecun-mnisthandwrittendigit-2010}.
    This dataset consists of images of the handwritten digits
    with labels
    from ``0" to ``9".
    We consider the following domain incremental learning setting \cite[Figure 1, Table 2]{VenTolias_2019}, \cite{ZenkePooleGanguli_2017}:  
    We split the data into tasks
    by separating the samples
    into the following $T=5$ tasks 
    (``0", ``1"),
    (``2", ``3"),
    (``4", ``5"),
    (``6", ``7"),
    (``8", ``9"),
    i.e., each of the $T=5$ tasks consists of samples of 
    one even digit and one odd digit.
    Hence, the continual learning task is to determine whether a given digit is even or odd when trained on sets of distinct pairs of even and odd digits. 

    We convert the $28\times 28$ pixel images,
    with pixel values on $[0,255]$,
    to $784\times 1$ vectors $\xvec_j\inrbb{784\times 1}$,
    with the values divided by $255$ such that entries of $\xvec_j$ lie on $[0,1]$.
    Using random features \cite{RahimiRecht_2008}
    the vectors $\xvec_j$ are transformed to regressors as
    $\avec_j = [\cos{}(\zetavec_1^T \xvec_j), \cdots, \cos{}(\zetavec_p^T \xvec_j)]\T\inrbb{p\times 1}$,
    where $\zetavec_\ell\inrbb{784\times 1}$ are i.i.d. random vectors with $\zetavec_\ell \sim \Nc(\bm 0, 0.04\eye{784})$.
    We use \cocoa{}  to train a model  $\what_{\text{odd},t}^{(T_c)}\inrbb{p\times 1}$ on the odd digits,
    and one model     $\what_{\text{even},t}^{(T_c)}\inrbb{p\times 1}$
    on the even digits in a continual learning setting.
    We apply the one-v.s.-rest classification strategy \cite{bishop2006pattern}
    to create predictions of the labels. We
     report the prediction error rate for each task 
    using  test sets of $2000$ samples unseen during training.
    \balert{%
    We use two different cost functions for training, the quadratic loss of \eqref{eqn:training_MSE},  and the logistic loss $\Ltrain{} (\what)  = \sum_{j=1}^{n}\ln(1+e^{-y_{j} \avec_{j}\T \what}) $, where $y_{j} \in \{-1,1\}$ is the label that corresponds to  $\avec_{j}$.}

    With $p=3000$,
    $n_t = 100$ ,
    $K=2$,
    $T_c = 1$,
    the $T=5$ tasks,
    and $100$ training repetitions over the task sequence $\trange$,
    we plot the prediction error rate per task 
    versus the training iteration index in Fig.~\ref{fig:mnist} \balert{for the quadratic loss and Fig.~\ref{fig:mnist:logistic} for the logistic loss.}
    Here,
    task 0 corresponds to (``0",``1"),
    task 1 to (``2", ``3"),
    etc.
     Although some of the tasks (e.g. Task $0$: ``0" v.s.``1") are giving significantly lower error rate   than others (e.g. Task 4: ``8" v.s.``9"),
     \cocoa{} is able to perform  significantly better than the random chance of $0.5$  for all tasks \balert{when trained with the quadratic loss or the logistic loss}.

%% file: conclusions.tex
\section{Discussions}\label{sec:discussions}

\balert{Our results show that generalization performance heavily depends on the number of nodes $K$. Hence,  one would want to optimize the performance of \cocoa{} by adjusting $K$,  but this may not be possible in all applications. In these cases, avoiding   the undesirable conditions mentioned in Remark~\ref{remark:localstability} and \eqref{eqn:condition_n_smaller_than_pk} -- \eqref{eqn:condition_n_larger_than_pk}  by assigning different number of parameters to each node,  using a subset of available nodes or  adjusting the model size  can be beneficial. Regularization can also be utilized  to prevent the high-error values at the interpolation threshold at the cost of slow convergence rate \cite[Fig.~8]{hellkvist_ozcelikkale_ahlen_linear_2021}. }

\balert{%
Our analytical characterizations focus either on the overparametrized local models (Assumption~\ref{assume:broad_and_large_Tc}) or the  one-shot setting (Assumption~\ref{assume:T_c_is_one}). On the other hand, an important trade-off within distributed learning is the one
between the number of communication rounds and the amount of local computations.
The \cocoa{} framework is flexible in terms of this trade-off
as it allows local solvers to be run to an arbitrary level of accuracy. Extending our analysis to explore the consequences of this flexibility for continual learning is a promising research direction.}

\section{Conclusions}\label{sec:conclusions}

We have focused on continual learning with the distributed optimization algorithm \cocoa{} and provided analytical expressions for the generalization error for a range of scenarios. These results revealed that under continual learning, the network structure may significantly affect the generalization error  in a manner that goes beyond what has been reported with only one task in distributed learning \cite{hellkvist_ozcelikkale_ahlen_linear_2021} and centralized continual learning \cite{lin_2023_theory}. We have quantified how the most favorable network structure for good generalization performance, such as  the number of nodes in the network, depends on the task similarity as well as  the number of tasks.

\galert{%
Characterizing the continual learning performance under general regressor models including correlation between features, different  communications schemes for \cocoa{}  and also for other distributed learning algorithms are considered  important directions for future work.
}%

%% file: appendix.tex
\appendix
\section{Appendix}

\subsection{Preliminaries}

The following lemma collects some properties of Gaussian random matrices that are used throughout the derivations: 

\begin{lemma}\label{lemma:useful_lemmas}
    If $\Amat_t = [\Amat_{t,[1]}, \cdots, \Amat_{t,[K]}]$,
    and $\Amat_{t,[k]}\inrbb{n_t\times p_k}$,
    with 
    $p_k \notin [n_t - 1,\, n_t + 1]$,
    and $p_k,\, n_t \geq 1$
    $\krange$,
    are all standard Gaussian random matrices,
    then
    \begin{align}\label{eqn:useful_ATA}
        & \eunder{\Amat_{t,[k]}}\left[\Amat_{t,[k]}\T \Amat_{t,[k]}\right]
        = n_t\eye{p_k},\\
        \label{eqn:useful_ApA}
        & \eunder{\Amat_{t,[k]}}\left[\Amat_{t,[k]}\p \Amat_{t,[k]}\right]
        = r_{t,k}\eye{p_k},\\
        & \eunder{\Amat_{t,[k]}}\left[\left(
            \Amat_{t,[k]} \Amat_{t,[k]}\T\right)\p\right]
        = \frac{\gamma_{t,k}}{n_t}\eye{n_t},\label{eqn:useful_AAT_p}\\
        & \eunder{\Amat_{t,[i]}\Amat_{t,[k]}}\!\!\left[
            \Amat_{t,[i]}\T\left(
            \Amat_{t,[k]} \Amat_{t,[k]}\T\right)\p\!\!
            \Amat_{t,[i]}
            \right]
        = \gamma_{t,k}\eye{p_i},\label{eqn:useful_ATAATpA}
    \end{align}
    with
    $ 
            r_{t,k} = \frac{ \min(n_t,p_k) }{p_k},\,
        \gamma_{t,k} = \frac{\min(n_t,p_k)}{\max(n_t,p_k) - \min(n_t,p_k) - 1} 
    $,
    and $i\neq k$.
    Additionally,
    if $\Hmat = \diag{h_k\eye{p_k}}{k=1}{K}\inrbb{p\times p}$ 
    is a diagonal matrix 
    with arbitrary $h_k\inrbb{}$, $\krange$,    
    then the following also hold,
    \begin{align}
        & \eunder{\Amat_t}
        \left[ \Hmat \Abar_t \Amat_t \right]
            = \diag{
                \frac{h_k r_{t,k}}{K} \eye{p_k}
            }{k=1}{K}\!, \label{eqn:expectation_AbartAt_isoG} \\
        & \eunder{\Amat_t}\left[\Hmat \Pmat_t\right]
            = \diag{
                h_k
                \left(1 - \frac{ r_{t,k}}{K} \right)\eye{p_k}
            }{k=1}{K}\!, \label{eqn:expectation_Pt_isoG} \\
        & \eunder{\Amat_t}\left[
                \Abar_t\T \Hmat \Abar_t
            \right]
             = \frac{1}{K^2}
             { \sum_{k=1}^K}
                    \frac{h_k \gamma_{t,k}}{n_t}\eye{n_t}, \label{eqn:expectation_AbarT_H_Abar}\\
        \begin{split}
        &  \eunder{\Amat_t}\!\left[\!
                \Amat_t\T \! \Abar_t\T \! \Hmat \! \Abar_t \Amat_t
            \right] \!
         \! = \! \diag{ \!\!
                \frac{h_k  r_{t,k} \! + \! 
                \sum_{\substack{i=1\\i\neq k}}^K \! h_i \gamma_{t,i}}{K^2}\eye{p_k}
             }{\!k=1}{\!K}\!\!\!\!, \label{eqn:expectation_ATAbarT_H_AbarA}
        \end{split}
        \\
        \begin{split}
        & \eunder{\Amat_t}\left[\Pmat_t\T \Hmat \Pmat_t\right] \label{eqn:PT_H_P}\\
        & \! = \! \diag{\!
            \frac{h_k (K^2 \!
                + \! r_{t,k}(1  \! - \! 2K)) \!
                + \! \sum_{\substack{i=1\\i\neq k}}^K \! h_i \gamma_{t,i}}{K^2}
            \eye{p_k}
             }{k=1}{K}\!\!\!. 
        \end{split}
    \end{align}
    
\end{lemma}

\noindent Proof:
The expression in \eqref{eqn:useful_ATA}
is a direct consequence of the fact that $\Amat_{t,[k]}\inrbb{n_t\times p_k}$ are standard Gaussian.
For \eqref{eqn:useful_ApA} see \cite[Eqn. (58)]{Hellkvist_FakeFeatures_2023_IEEETSP},
and note that for $p_k < n_t$,  $\Amat_{t,[k]}$ is full column-rank and $\Amat_{t,[k]}\p\Amat_{t,[k]}=\eye{p_k}$.
For \eqref{eqn:useful_AAT_p}, see \cite{cook_forzani_wishart_2011}.
Combining \eqref{eqn:useful_ATA} and \eqref{eqn:useful_AAT_p},
one obtains \eqref{eqn:useful_ATAATpA}.
We obtain \eqref{eqn:expectation_AbartAt_isoG} -- \eqref{eqn:PT_H_P}
 from 
\eqref{eqn:useful_ATA} -- \eqref{eqn:useful_ATAATpA} 
with algebraic manipulations together with the definitions of $\Pmat_t$ and $\Abar_t$ in Lemma~\ref{lemma:recursion}.
\qed

The following lemma gives an expression for the average distance of the 
\cocoa{} solution 
$\what_t^{(T_c)}$
to a given vector
$\uvec$
for an arbitrary distribution for $\Amat_i$. 
We use a version of this result specialized to Gaussian case in the proof of Theorem~\ref{thm:expected_gen_isoG}.  

\begin{lemma}\label{lemma:recursion_sum_of_alphas}
    Let $\Amat_1,\, \dots,\, \Amat_t, \, \zvec_1, \, \dots,\, \zvec_t$
    be uncorrelated
    and the noise vectors $\zvec_\tau$ be zero-mean.
    If all partitions $\Amat_{\tau,[k]}$,
    $\tau=1,\,\dots,\,t,~\krange$,
    are full row rank or $T_c=1$,
    then for any fixed $\uvec\inrbb{p\times 1}$,
    \begin{equation}\label{eqn:general_wt-wi_Tc_1}
        \eunder{\traindata{t}}\left[\left\|\what_t^{(T_c)} - \uvec\right\|^2 \right]
        = \left\|\uvec\right\|_{\Hmatt{1}{t}}^2
        + \sum_{\tau=1}^t \alpha_\tau,
    \end{equation}
    where
    \begin{align}\label{eqn:alpha_tau_final_Pl_vl}
    \begin{split}
        & \alpha_\tau 
         = \! \left\|\wvec_\tau^* - \uvec\right\|^2_{
            \eunder{\Amat_\tau}\left[
                \Amat_\tau\T\Abar_\tau\T\Hmatt{\tauplusone}{t}\Abar_\tau\Amat_\tau
            \right]
            } \\
        & \qquad\quad   + \! \eunder{\zvec_\tau}\left[
            \left\|\zvec_\tau\right\|^2_{
            \eunder{\Amat_\tau}\left[
                \Abar_\tau\T\Hmatt{\tauplusone}{t} \Abar_\tau
            \right]}
        \right] 
        + \! 2 \Bigg\langle\!
            \wvec_\tau^* \! - \! \uvec, \! \\
        & \qquad\quad
        \sum_{j=0}^{\tau-1} 
        \left(
            \prod_{\ell=j+1}^{\tau-1}\!\!
            \eunder{\Amat_\ell}\left[ \Pmat_\ell\right]
        \right)
        \eunder{\Amat_j}\left[
            \svec_j
        \right]
        \Bigg\rangle_{\!\!\!\!
        \eunder{\Amat_\tau}\left[
            \Amat_\tau\T\Abar_\tau\T \Hmatt{\tauplusone}{t} \Pmat_\tau
        \right]},
    \end{split}
    \end{align}
    where $\Abar_\tau$ and $\Pmat_\tau$ are defined in Lemma~\ref{lemma:recursion},
    $\Hmatt{t+1}{t} = \eye p$,
    and 
    \begin{equation}
        \Hmatt{\tau}{t}
        = \Ebb_{\Amat_\tau,\dots,\Amat_t}[
            \Pmat_\tau\T \cdots \Pmat_t\T
            \Pmat_t \cdots \Pmat_\tau
        ],~~
        \tau = 1,\,\dots,\, t,
    \end{equation}
    and where $\svec_0 = -\uvec$,
    and 
    \begin{equation}
        \svec_j = \Abar_j\Amat_j(\wvec_j^* - \uvec),~~
        j = 1, \, \dots, \, t - 1. 
    \end{equation}
    
\end{lemma}

Proof: See Appendix~\ref{proof:lemma:recursion_sum_of_alphas}.

\subsection{Proof of Lemma~\ref{lemma:recursion}}
\label{proof:lemma:recursion}

        From Algorithm~\ref{alg:continual_cocoa},
    $
        \vvec_{t,[k]}^{(0)} \!=\! K \Amat_{t,[k]} \what_{t,[k]}^{(0)}, 
    $
    and
    $
        \vbar_t^{(1)} \!=\! \frac{1}{K}\sum_{k=1}^K\vvec_{t,[k]}^{(0)}
        = \Amat_t \what_t^{(0)}.    
    $
    Then,
    $
        \Delta \what_{t,[k]}^{(1)}
        \!=\! - \frac{1}{K}\Amat_{t,[k]}\p \Amat_t \what_t^{(0)} + \frac{1}{K}\Amat_{t,[k]}\p \yvec_t,
    $
    $
        \what_{t,[k]}^{(1)}
        \!=\! \what_{t,[k]}^{(0)} - \frac{1}{K}\Amat_{t,[k]}\p \Amat_t \what_t^{(0)} + \frac{1}{K}\Amat_{t,[k]}\p \yvec_t.
    $
    Stacking $\what_{t,[k]}^{(1)}$,
    as in \eqref{eqn:wvec_k_stacked} with $i=1$,
    \begin{equation}
        \what_{t}^{(1)}\!\!
        = 
        \what_t^{(0)} 
        - \frac{1}{K} \!\!
            \begin{bmatrix}
                \Amat_{t,[1]}\p \\\vdots\\
                \Amat_{t,[K]}\p
            \end{bmatrix}\!
            \Amat_t \what_t^{(0)} 
        + \frac{1}{K}\!\!
            \begin{bmatrix}
                \Amat_{t,[1]}\p \\\vdots\\
                \Amat_{t,[K]}\p
            \end{bmatrix}
            \yvec_t.
    \end{equation}
    Inserting the initialization $\what_t^{(0)} = \what_{t-1}^{(T_c)}$,
    we obtain \eqref{eqn:recursion}.

    The expression in \eqref{eqn:recursion_minus_wi} is obtained
    by adding and subtracting 
    $\Abar_t\Amat_t\uvec$ to $\what_t^{(1)} - \uvec$,
    and opening up $\yvec_t = \Amat_t \wvec_t^* + \zvec_t$ and re-arranging. 
    This concludes the proof.

\subsection{Proof of Theorem~\ref{thm:expected_gen_isoG}}
\label{proof:thm:expected_gen_isoG}
The expected generalization error,
defined in \eqref{eqn:expected_gen_err_def},
 is
\begin{equation}
    G_T(\what_{t}^{(T_c)})
    = \frac{1}{T}\sum_{i=1}^T
    \eunder{\traindata{t}}
    \left[ \|\what_{t}^{(T_c)} - \wvec_i^*\|^2\right] + \sigma_i^2.
\end{equation}
We will show that,
with the definitions given in Theorem~\ref{thm:expected_gen_isoG},
\begin{align}\label{eqn:wt_minus_wi_final_lemma}
    \eunder{\traindata{t}}\left[
        \left\|\what_t^{(T_c)} \! - \! \wvec_i^*\right\|^2
    \right]
    & = \left\|\wvec_i^*\right\|^2_{\Hmatt{1}{t}}
    + \phi\left(\what_t^{(T_c)}, \wvec_i^*\right).
\end{align}
Lemma~\ref{lemma:recursion_sum_of_alphas} gives this expression up to the expectations over the distributions of regressors and noise vector,
with 
\begin{equation}\label{eqn:phi_sum_of_alphas}
    \phi\left(\what_t^{(T_c)}, \wvec_i^*\right) = \sum_{\tau=1}^t \alpha_\tau,
\end{equation}
where we use $\uvec=\wvec_i^*$ in $\alpha_\tau$.

We now continue with deriving the expectations of the expressions in Lemma~\ref{lemma:recursion_sum_of_alphas},
in the setting of Theorem~\ref{thm:expected_gen_isoG}.

The matrices $\Pmat_\tau=\eye{p} - \Abar_\tau\Amat_\tau$, $1 \leq \tau \leq  t$, 
are uncorrelated and
we have 
$\Hmatt{1}{t}
    = \eunder{\traindata{t-1}}\left[
        \Pmat_1\T \cdots \Pmat_{t-1}\T \Hmatt{t}{t} \Pmat_{t-1} \cdots \Pmat_1
    \right]$,
with $\Hmatt{t}{t} = \eunder{\Amat_t}\left[\Pmat_t\T\Pmat_t\right] = \diag{h_{t,k} \eye{p_k}}{k=1}{K}$,
where
\begin{equation}
   h_{t,k}
    = \frac{
        K^2 + r_{t,k}(1-2K) + \sum_{\substack{i=1\\i\neq k}}^K \gamma_{t,i}
    }{K^2},
\end{equation}  
by \eqref{eqn:PT_H_P}. 
We can repeat this for 
$\Hmatt{t-1}{t}
    = \eunder{\Amat_{t-1}}\left[
        \Pmat_{t-1}\T \Hmatt{t}{t} \Pmat_{t-1}
    \right]
    = \diag{h_{t-1,k} \eye{p_k}}{k=1}{K}$,
where
\begin{equation}
    h_{t-1,k} \!
    = \! \frac{\! 
        h_{t,k}(K^2 \! + \! r_{t-1,k}(1\!-\!2K)) 
        \!+\! \sum_{\substack{i=1\\i\neq k}}^K h_{t,i} \gamma_{t-1,i}
    }{K^2}.
\end{equation}
Repeating until 
$\Hmatt{1}{t} = \eunder{\Amat_1}\left[
        \Pmat_1\T \Hmatt{2}{t} \Pmat_1
    \right]$
gives the desired expression in \eqref{eqn:h_tau_k_main_thm}.
By combining this with \eqref{eqn:general_wt-wi_Tc_1},
we obtain the desired expression for the first term in 
\eqref{eqn:wt_minus_wi_final_lemma}.

We now continue with the second term in \eqref{eqn:wt_minus_wi_final_lemma}.
With $\Hmatt{\tauplusone}{t} = \diag{h_{\tau+1,k}\eye{p_k}}{k=1}{K}$,
we apply \eqref{eqn:expectation_ATAbarT_H_AbarA}
to write
\begin{align}
\begin{split}\label{eqn:AT_AbarT_H_Abar_A_proof}
    & \eunder{\Amat_\tau}\left[
        \Amat_\tau\T\Abar_\tau\T
        \Hmatt{\tauplusone}{t}
        \Abar_\tau\Amat_\tau
    \right] \\
    & \qquad = \diag{
        \frac{
            h_{\tauplusone} r_{\tau,k}
            + \sum_{\substack{i=1\\i\neq k}}^K
                h_{\tauplusone,i}\gamma_{\tau,i} }
        {K^2} \eye{p_K}
    }{k=1}{K}
\end{split},
\end{align}
and use \eqref{eqn:expectation_AbarT_H_Abar}
to write
\begin{equation}
    \eunder{\Amat_\tau}\left[
        \Abar_\tau\T
        \Hmatt{\tauplusone}{t}
        \Abar_\tau
    \right]
    = \frac{1}{K^2}\sum_{k=1}^K
        \frac{h_{\tauplusone,k} \gamma_{\tau,k}}{n_\tau}\eye{n_\tau}.
\end{equation}
Here,
$\zvec_\tau \sim \Nc(\bm 0, \sigma_\tau^2 \eye{n_\tau})$,
hence
$\Ebb_{\zvec_\tau}[\|\zvec_\tau\|^2] = n_\tau \sigma_\tau^2$,
and 
\begin{equation}\label{eqn:noise_term_proof}
    \eunder{\zvec_\tau}\left[
        \left\|\zvec_\tau\right\|^2_{
        \eunder{\Amat_\tau}\left[
            \Abar_\tau\T
            \Hmatt{\tauplusone}{t}
            \Abar_\tau
        \right]}
    \right]
    = \sigma_\tau^2 
        \frac{\sum_{k=1}^K h_{\tauplusone,k} \gamma_{\tau,k}}{K^2}.
\end{equation}
Combining \eqref{eqn:expectation_AbartAt_isoG} and \eqref{eqn:expectation_Pt_isoG},
we find that for $j \geq 1$,
\begin{align}
\begin{split}
    &\prod_{\ell=j+1}^{\tau-1}\!\!
            \eunder{\Amat_\ell}\left[ \Pmat_\ell\right]
            \eunder{\Amat_j}\left[\Abar_j \Amat_j \right] \\
    & = \prod_{\ell=j+1}^{\tau-1}\!\!
    \diag{\left(1 - \frac{r_{\ell,k}}{K}\right)\eye{p_k}}{k=1}{K}
    \diag{\frac{r_{j,k}}{K}\eye{p_k}}{k=1}{K}
\end{split}\\
& = \diag{
    \frac{r_{j,k}}{K}
    \prod_{\ell=j+1}^{\tau-1}
    \left(1 - \frac{r_{\ell,k}}{K}\right)
    \eye{p_k}
    }{k=1}{K},
    \label{eqn:prod_Pl_AbarA_j}
\end{align}
and for $j=0$,
\begin{align}\label{eqn:prod_Pl}
    \prod_{\ell=1}^{\tau-1}
        \eunder{\Amat_\ell}\left[ \Pmat_\ell\right]
    & = \diag{
        \prod_{\ell=1}^{\tau-1}
            \left(1 - \frac{r_{\ell,k}}{K}\right)
        \eye{p_k}
    }{k=1}{K}.
\end{align}

Using 
\eqref{eqn:expectation_AbartAt_isoG}
and \eqref{eqn:expectation_ATAbarT_H_AbarA},
we have that
\begin{align}\begin{split}
    & \eunder{\Amat_\tau}\left[
        \Amat_\tau\T\Abar_\tau\T
        \Hmatt{\tauplusone}{t} 
        \Pmat_\tau
    \right] \\
    & =  \eunder{\Amat_\tau}\left[
            \Amat_\tau\T\Abar_\tau\T
            \Hmatt{\tauplusone}{t}
        \right]
    - \eunder{\Amat_\tau}\left[
            \Amat_\tau\T\Abar_\tau\T
            \Hmatt{\tauplusone}{t}
            \Abar_\tau \Amat_\tau 
        \right]
\end{split} \\
\begin{split}\label{eqn:Exp_AAbarT_H_P}
    & = \diag{\frac{h_{\tauplusone,k} r_{\tau,k}}{K} \eye{p_k} }{k=1}{K} \\
    & \qquad - \diag{\frac{h_{\tauplusone,k} r_{\tau,k}
        + \sum_{\substack{i=1\\i\neq k}}^K
        h_{\tauplusone,i} \gamma_{\tau,i}}{K^2} \eye{p_k}}{k=1}{K}.
\end{split}
\end{align}

We now combine 
\eqref{eqn:AT_AbarT_H_Abar_A_proof},
\eqref{eqn:noise_term_proof},
\eqref{eqn:prod_Pl_AbarA_j},
\eqref{eqn:prod_Pl}
and \eqref{eqn:Exp_AAbarT_H_P},
we find the final form of $\alpha_\tau$ in \eqref{eqn:phi_sum_of_alphas},
\begin{align}
\begin{split}
    \alpha_\tau 
     & = \left\|\wvec_\tau^* - \wvec_i^*\right\|_{\Rmatt{\tau}{t}}^2
    + \sigma_\tau^2 
        \frac{\sum_{k=1}^K h_{\tauplusone,k} \gamma_{\tau,k}}{K^2} \\
    & ~\quad\qquad  
        + 2 \sum_{j=0}^{\tau - 1}
        \left\langle
            \wvec_\tau^* - \wvec_i^*, \wvec_j^* - \wvec_i^*
        \right\rangle_{\Qmatt{\tau}{j}{t}},
\end{split}
\end{align}
with $\Rmatt{\tau}{t}$ and $\Qmatt{\tau}{j}{t}$ defined as in Theorem~\ref{thm:expected_gen_isoG}.
Note that we set $\wvec_0^* = \bm 0$ and $r_{0,k} = K$
to make the notation more compact.
By inserting this expression for $\alpha_\tau$ together with the derived expression for $\Hmatt{1}{t}$ into \eqref{eqn:general_wt-wi_Tc_1},
we obtain the desired expression in \eqref{eqn:wt_minus_wi_final_lemma},
concluding the proof of Theorem~\ref{thm:expected_gen_isoG}.

\subsection{Proof of Theorem~\ref{thm:similarity_generr}}
\label{proof:thm:similarity_generr}
    By inserting $\wvec_t^*=\wvec^*$ and $\sigma_t^2 = \sigma^2$
    into $G_T(\what_T^{(T_c)})$ in \eqref{eqn:gen_err_isoG}
    and simplifying,
    the desired expression in \eqref{eqn:gen_err_similar_tasks} is obtained.

    We now prove \eqref{eqn:condition_n_smaller_than_pk} and \eqref{eqn:condition_n_larger_than_pk}.
    Recall the definition 
    $\Hmatt{1}{T} = \diag{h_{1,k}\eye{p_k}}{k=1}{K}$,
    with 
    \begin{align}\label{eqn:}
        h_{\tau,k} \! = \! \frac{
            h_{\tauplusone,k}\left(K^2 \! + \! r_{\tau,k} (1 \! - \! 2K)\right) 
            + \sum_{\substack{i=1\\i\neq k}}^K 
                h_{\tauplusone,i} \gamma_{\tau,i} 
        }{K^2},
    \end{align}
    $\tau = 1,\dots,T$,
    and $h_{\Tplusone,k} = 1$, $\krange$.
    Note that $K^2 + r_{\tau,k}(1 - 2K) \geq 0$
    and $\gamma_{\tau,k} > 0$,
    hence the fraction here is non-negative,
    and
    \begin{equation}
        h_{\tau,k} \leq h_{\tauplusone,\max}
            \frac{
                K^2 - r_{\tau,k} ( 2K -1 )
                + \sum_{\substack{i=1\\i\neq k}}^K \gamma_{\tau,i} 
            }{K^2},
    \end{equation}
    where $h_{\tau,\max} = \max_{k} h_{\tau,k}$.
    Note that this upper bound is bounded above by replacing 
    $r_{\tau,k}$ with $r_{\tau,\min}=\min_{k} r_{\tau,k}$
    and $\gamma_{\tau,k}$ with $\gamma_{\tau,\max} = \max_k \gamma_{\tau,k}$,
    hence
    \begin{equation}\label{ineq:h_tau_htauplus1_general}
        h_{\tau,k} \leq h_{\tauplusone,\max} \times 
            f_\tau,
    \end{equation}
    where 
    \begin{equation}
        f_\tau = \frac{K^2 -  (2K - 1) r_{\tau,\min} + (K-1)\gamma_{\tau,\max} }{K^2}.
    \end{equation}
    We have that
    \begin{equation}
        h_{1,k} \leq h_{2,\max} f_1 \leq \dots \leq f_T \cdots f_1,
    \end{equation}
    where we have used that $h_{\Tplusone,k} = 1$.
    
    It follows that if $|f_\tau| < 1,\,\forall\tau$,
    then
    $
        \lim_{T\to\infty} h_{1,k} = 0,
    $
    and hence
    $
        \lim_{T\to\infty} \|\wvec^*\|_{\Hmatt{1}{T}}^2 = 0.
    $
    We recall that the fraction $f_\tau$ is nonnegative and continue by deriving the  conditions for which $f_\tau < 1$. 
    We will consider the following two cases separately:

    \paragraph{$n_\tau < p_k, ~\forall \tau, k$}
    Here, 
    $r_{\tau,\min} = \frac{n_\tau}{\pmax}$
    and $\gamma_{\tau,\max} = \frac{n_\tau}{\pmin - n_\tau - 1}$.
    Inserting these identities into 
    $f_\tau < 1$,
    we  obtain 
    \begin{align}
        \frac{
            K^2 
            - (2K - 1) \frac{n_\tau}{\pmax} 
            + (K-1)\frac{n_\tau}{\pmin - n_\tau - 1}
        }{K^2}
        < 1. 
    \end{align}
    Simplifying this expression gives the condition in \eqref{eqn:condition_n_smaller_than_pk}.
    Note that the bound in \eqref{eqn:condition_n_smaller_than_pk} is strictly less than $ \pmin$,
    hence the condition $n_\tau < p_k$ of the setting here is included in this bound.

    \paragraph{$n_\tau > p_k,~\forall \tau, k$}
    Here, 
    $r_{\tau,\min} = 1$
    and $\gamma_{\tau,\max} = \frac{\pmin}{n_\tau - \pmax - 1}$.
    Setting $f_\tau < 1$,
    we obtain 
    \begin{align}
         \frac{
            K^2 
            - (2K - 1)
            + (K-1)\frac{\pmin}{n_\tau - \pmax - 1}
        }{K^2}
        < 1. 
    \end{align}
    Re-arranging this expression gives the bound in \eqref{eqn:condition_n_larger_than_pk}.
    Note that the bound in \eqref{eqn:condition_n_larger_than_pk} is  strictly greater than $ \pmax$,
    hence $n_\tau > p_k$ is fulfilled if the bound holds.
    This concludes the proof.

\subsection{Proof of Lemma~\ref{col:forgetting_similarity}}
\label{proof:col:forgetting_similarity}
    Inserting $\wvec_t^*=\wvec^*$ and $\sigma_t^2 = 0$
    into $F_T(\what_T^{(T_c)})$ in \eqref{eqn:expected_forgetting},
    \begin{align}
        F_T(\what_T^{(T_c)}) 
        & = \frac{1}{T}\sum_{i=1}^T \frac{1}{2 n_i} 
        \eunder{\traindata{T}}\left[
            \left\|
                \Amat_i (\what_T^{(T_c)} - \wvec^*)
            \right\|^2
        \right].
    \end{align}
    We apply the submultiplicativity of the $\ell_2$-norm,
    \begin{align}\label{eqn:forgetting_bound_col}
        F_T(\what_T^{(T_c)}) 
        &  \leq \frac{1}{T}\sum_{i=1}^T \frac{1}{2 n_i} 
        \eunder{\traindata{T}}\left[
            \left\| \Amat_i \right\|^2 
            \left\|
                \what_T^{(T_c)} - \wvec^*
            \right\|^2
        \right].
    \end{align}
    For a large number of tasks $T$, hence for a large number of i.i.d. samples, 
    the estimate $\what_T^{(T_c)}$ may be assumed to be uncorrelated with $\Amat_i$ \cite[Ch. 16]{b_HaykinAdaptive}. 
    Hence, we approximate
    \begin{equation}\label{eqn:indAhatw}
    \!\eunder{\traindata{T}}\!\!\left[\!
            \left\| \Amat_i \right\|^2 \!
            \left\|
                \what_T^{(T_c)} \! \! - \! \wvec^*
            \right\|^2
        \right] \!\! \approx \!\!
        \eunder{\Amat_i}\!\left[
            \left\| \Amat_i \right\|^2
        \right] \!\!
        \eunder{\traindata{T}}\!\left[
            \left\|
                \what_T^{(T_c)} \!\! - \! \wvec^*
            \right\|^2
        \right]\!.
    \end{equation}
    From the proof of Theorem~\ref{thm:expected_gen_isoG},
    we have \eqref{eqn:wt_minus_wi_final_lemma},
    which here becomes
    \begin{equation}\label{eqn:wT-w_star_col}
        \eunder{\traindata{T}}\left[
        \left\|
            \what_T^{(T_c)} - \wvec^*
        \right\|^2 \right]
        = \|\wvec^*\|_{\Hmatt{1}{T}}^2,
    \end{equation}
    where we have used that $\phi\!\left(\what_t^{(T_c)}\!, \wvec_i^*\right) = 0$, $\tau=1,\,\dots,\, T$,
    because $\wvec_\tau^* = \wvec^*$ and $\sigma_\tau^2 = 0$, see for instance \eqref{eqn:phi:result}.
    Now if either the condition in 
    \eqref{eqn:condition_n_smaller_than_pk} or 
    \eqref{eqn:condition_n_larger_than_pk} holds for $t=1,\dots,T$,
    then we can apply \eqref{eqn:limit_H1_term} to \eqref{eqn:wT-w_star_col}, to obtain
    $
        \lim_{T\to\infty}
        \eunder{\traindata{T}}\left[
        \left\|
            \what_T^{(T_c)} - \wvec^*
        \right\|^2 \right]
        = 0.
    $
    Inserting this into \eqref{eqn:indAhatw}, and \eqref{eqn:indAhatw} into \eqref{eqn:forgetting_bound_col}, we find that 
    the upper bound in \eqref{eqn:forgetting_bound_col} is approximately zero in the limit of $T\to\infty$,
    i.e.,
    $
        \lim_{T\to\infty} F_T\left(\what_T^{(T_c)}\right)  \approx 0.
    $
    This concludes the proof.

\subsection{Proof of Lemma~\ref{lemma:recursion_sum_of_alphas}}
\label{proof:lemma:recursion_sum_of_alphas}

We first give 
the following lemma which gives the expression for the last step of the recursion.
\begin{lemma}\label{lemma:first_recusion_variance}
    Let $T_c=1$
    or the partitions of the latest task,
    task $\tau$,
    $\Amat_{\tau,[k]}$ be full row rank.
    Also let
    the regressors in $\Amat_\tau$ be uncorrelated with 
    the zero-mean noise vector $\zvec_\tau$
    and the previous tasks.
    Then,
    for any fixed $\uvec\inrbb{p\times 1}$
    and 
    %
    %
    some symmetric matrix
    $\Hmat\inrbb{p\times p}$,
    we have
    \begin{align}\label{eqn:norm+alpha}
        & \eunder{\traindata{\tau}}\!\left[\!
            \left\|\what_\tau^{(T_c)} - \uvec \right\|^2_{\Hmat}
        \right]
        \! = \! \eunder{\traindata{\tau-1}}\!\left[\!
            \left\|\what_{\tau-1}^{(T_c)} - \uvec \right\|^2_{\Hmat_{\tau}}
        \right] 
        + \alpha_\tau,
    \end{align}
    where
    $\Hmat_{\tau} 
        = \eunder{\Amat_\tau}\left[
            \Pmat_\tau\T\Hmat \Pmat_\tau
        \right]$
    and
    \begin{align}
       \nonumber
        \alpha_\tau 
        & = \! \left\|\wvec_\tau^* - \uvec\right\|^2_{
            \eunder{\Amat_\tau}\left[
                \Amat_\tau\T\Abar_\tau\T\Hmat\Abar_\tau\Amat_\tau
            \right]
            }
         + \! \eunder{\zvec_\tau}\left[
            \left\|\zvec_\tau\right\|^2_{
            \eunder{\Amat_\tau}\left[
                \Abar_\tau\T\Hmat \Abar_\tau
            \right]}
        \right] \\
        \label{eqn:alpha_tau_def}
        & \quad  + \! 2 \left\langle\!
            \wvec_\tau^* \! - \! \uvec, \! 
            \eunder{\traindata{\tau-1}}\!\!\left[\what_{\tau-1}^{(T_c)} \! - \! \uvec\right] 
        \right\rangle_{\!\!\!\!
        \eunder{\Amat_\tau}\left[
            \Amat_\tau\T\Abar_\tau\T\Hmat \Pmat_\tau
        \right]}.
    \end{align}

\end{lemma}
Proof: See Appendix~\ref{proof:lemma:first_recusion_variance}.

We will now use \eqref{eqn:norm+alpha}  on $\eunder{\traindata{t}}\left[\left\|\what_t^{(T_c)} - \uvec\right\|^2 \right]$ recursively. 
Starting with $\Hmat = \eye{p}$, we
 write
\begin{align}
   \nonumber
    & \eunder{\traindata{t}}\left[\left\|\what_t^{(T_c)} - \uvec\right\|^2 \right]
    = \! \eunder{\traindata{t-1}}\!\left[\!
            \left\|\what_{t-1}^{(T_c)} - \uvec \right\|^2_{\eunder{\traindata{t}}\left[
            \Pmat_t\T \Pmat_t
        \right]}
        \right] 
        + \alpha_t \\
    & = \! \eunder{\traindata{t-2}}\!\left[\!
            \left\|\what_{t-2}^{(T_c)} - \uvec \right\|^2_{\eunder{\traindata{t}}\left[
            \Pmat_{t-1}\T \Pmat_t\T \Pmat_t \Pmat_{t-1}
        \right]}
        \right]
        + \alpha_{t} + \alpha_{t-1} \\
    & = \dots 
    = \left\|\uvec \right\|^2_{\Hmatt{1}{t}}
        + {\textstyle \sum_{\tau=1}^t } \alpha_{\tau},
        \label{eqn:alpha_proof_insert_bias}
\end{align}
where 
$\Hmatt{1}{t} = \eunder{\traindata{t}}\left[
            \Pmat_{1}\T \cdots \Pmat_t\T \Pmat_t \cdots \Pmat_{1}
        \right]$
and 
we have used that we initialize the continual learning procedure with 
$\wvec_0 = \bm 0$.

\begin{lemma}\label{lemma:expected_bias_general}
    Within the setting of Lemma~\ref{lemma:recursion_sum_of_alphas},
    the following holds,
    \begin{align}\label{eqn:expected_bias}
        \eunder{\traindata{\tau-1}}\left[
            \what_{\tau-1}^{(T_c)} - \uvec
        \right]
        & = \sum_{j=0}^{\tau-1}\!\left(\!
            \prod_{\ell=j+1}^{\tau-1}\!\!\left(
            \eunder{\Amat_\ell}\left[ \Pmat_\ell\right]\!\right)\!\!
            \eunder{\Amat_j, \zvec_j}[\svec_j]\!\!
        \right),
    \end{align}
    where $\svec_0 = -\uvec$
    and 
    $\svec_j
        = \Abar_j \Amat_j 
                (\wvec_j^* - \uvec) 
            + \Abar_j \zvec_j$,
    $j=1,\,\dots,\,\tau-1$.
\end{lemma}

Proof: See Appendix~\ref{proof:lemma:expected_bias_general}.

Inserting the result of Lemma~\ref{lemma:expected_bias_general} into 
\eqref{eqn:alpha_tau_def}
and combining with \eqref{eqn:alpha_proof_insert_bias},
the desired expression for $\alpha_\tau$ is obtained.

\subsection{Proof of Lemma~\ref{lemma:first_recusion_variance}}
\label{proof:lemma:first_recusion_variance}
Using the recursion in \eqref{eqn:recursion_minus_wi},
and the notation $\wvec_\tau = \what_\tau^{(T_c)}$
and $\wvec_{\tau-1} = \what_{\tau-1}^{(T_c)}$,
we can write
\begin{align}
    \nonumber
    & \left\|
        \wvec_\tau - \uvec
    \right\|_{\Hmat}^2
    = 
    \left\|\Pmat_\tau (\wvec_{\tau-1} - \uvec)\right\|^2_{\Hmat} 
      +\! \left\|\Abar_\tau \Amat_\tau (\wvec_\tau^* - \uvec)\right\|^2_{\Hmat} \\
     \nonumber
    & + \left\| \Abar_\tau \zvec_\tau \right\|^2_{\Hmat} 
      +\! 2 \left\langle 
        \Abar_\tau \Amat_\tau (\wvec_\tau^* - \uvec),
        \Pmat_\tau (\wvec_{\tau-1} - \uvec)
    \right\rangle_{\Hmat} \\
    & ~ +\! 2 \left\langle 
        \Pmat_\tau (\wvec_{\tau-1}\! -\! \uvec) \!
        + \! \Abar_\tau \Amat_\tau (\wvec_\tau^* \! - \! \uvec),
        \Abar_\tau \zvec_\tau
    \right\rangle_{\Hmat}
\end{align}
Since $\Hmat$ is fixed,
hence independent from the random entities,
the expectation w.r.t. $\traindata{\tau}$ of
the final term is zero,
because $\zvec_\tau$ is zero-mean and 
uncorrelated with 
$\wvec_{\tau-1}$ and $\Amat_\tau$,
hence also with $\Pmat_\tau$.
For the other terms,
the desired expression is obtained by using the uncorrelatedness between 
$\wvec_{\tau-1}$ and $\Amat_\tau$,
and between $\zvec_\tau$ and $\Amat_\tau$.
We here illustrate the derivation of the first term,
\begin{align*}\begin{split}
    & \Ebb_{\traindata{\tau}}\left[
        \left\|\Pmat_\tau (\wvec_{\tau-1} - \uvec)\right\|^2_{\Hmat}
    \right] \\
    & = \Ebb_{\traindata{\tau-1}}\left[
         (\wvec_{\tau-1} - \uvec)\T 
         \Ebb_{\Amat_\tau}\left[\Pmat_\tau\T
         \Hmat
         \Pmat_\tau\right]
         (\wvec_{\tau-1} - \uvec)
    \right]
\end{split} \\
    & = \Ebb_{\traindata{\tau-1}}\left[\left\|
        \wvec_{\tau-1} - \uvec\right\|^2_{
        \Ebb_{\Amat_\tau}\left[\Pmat_\tau\T
         \Hmat
         \Pmat_\tau\right]
        }            
     \right]. 
\end{align*}
The remaining terms are found in a similar fashion.

\subsection{Proof of Lemma~\ref{lemma:expected_bias_general}}
\label{proof:lemma:expected_bias_general}
Here, we use the notation notation $\wvec_j = \what_j^{(T_c)}$,
for $j=0,\,\dots,\,\tau-1$.
Since the recursion in \eqref{eqn:recursion_minus_wi} holds for  
the tasks $1,\,\dots,\,t-1$,
we have
\begin{align}
    & \wvec_{\tau-1} - \uvec
    = \Pmat_{\tau-1}(\wvec_{\tau-2} - \uvec)
    + \svec_{\tau-1} \\
    & \qquad = \Pmat_{\tau-1}(\Pmat_{\tau-2}(\wvec_{\tau-3} - \uvec) + \svec_{\tau-2})
    + \svec_{\tau-1} \\
    \begin{split}
    & \qquad = \Pmat_{\tau-1} \cdots \Pmat_1 (\wvec_0 - \uvec) \\
    & \qquad \quad + \Pmat_{\tau-1} \cdots \Pmat_2 \svec_{1}
        + \cdots 
        + \Pmat_{\tau-1} \svec_{\tau-2}
        + \svec_{\tau-1}
    \end{split} \\
    & \qquad  
        = {\textstyle \sum_{j=0}^{\tau-1} }
        \left(
            \textstyle{\prod_{\ell=j+1}^{\tau-1}}
            \Pmat_\ell
        \right)
            \svec_j,
\end{align}
where $\svec_0 = -\uvec$.
The matrices $\Amat_\ell$, $\ell=1,\,\dots,\,t-1$ are all uncorrelated
with each other and with $\zvec_\ell$,
thus taking the expectation w.r.t. $\traindata{\tau-1}$ 
gives the desired expression.